%% file: main.tex
\documentclass[10pt,journal,compsoc]{IEEEtran}
\ifCLASSOPTIONcompsoc
  \usepackage[nocompress]{cite}
\else
  \usepackage{cite}
\fi

\usepackage{amsmath}
\usepackage{amssymb}
\usepackage{array}
\usepackage{bbm}
\usepackage{calc}
\usepackage{caption}
\usepackage{color}
\usepackage{xcolor}
\usepackage{enumitem}
\usepackage{epsfig}
\usepackage{epstopdf}
\usepackage{hyperref}
\usepackage{multirow}
\usepackage{subfigure}
\usepackage{booktabs}
\usepackage{url}
\usepackage{xspace}
\usepackage{pdfpages}
\usepackage{graphicx}
\usepackage{url}
\newcommand{\red}[1]{\textcolor{black}{#1}}

\def\onedot{\ifx\@let@token.\else.\null\fi\xspace}

\begin{document}
%\includepdf[pages=-,pagecommand={},width=\textwidth]{cover_letter.pdf}

% \begin{figure*}[hbtp!]
% \includegraphics[page=1,width=\textwidth]{cover_letter.pdf}
% \end{figure*}

\title{On the Importance of Visual Context for Data Augmentation in Scene Understanding}
\author{Nikita~Dvornik,
        Julien~Mairal,~\IEEEmembership{Senior~Member,~IEEE,}
        and~Cordelia~Schmid,~\IEEEmembership{Fellow,~IEEE}%
\IEEEcompsocitemizethanks{\IEEEcompsocthanksitem The authors are with University 
Grenoble Alpes, Inria, CNRS, Grenoble INP, LJK, 38000 Grenoble, France.
\protect\\
E-mail: firstname.lastname@inria.fr}
}

\IEEEtitleabstractindextext{%
\begin{abstract}
  \input{abstract.tex}
\end{abstract}

\begin{IEEEkeywords}
Convolutional Neural Networks, Data Augmentation, Visual Context, Object Detection, Semantic Segmentation. 
\end{IEEEkeywords}}

\maketitle

\IEEEdisplaynontitleabstractindextext
\IEEEpeerreviewmaketitle

\IEEEraisesectionheading{\section{Introduction}\label{sec:introduction}}

\input{intro.tex}

\section{Related Work}
\input{related.tex}

\section{Approach}
\input{approach.tex}

\section{Experiments}\label{sec:exp}
\input{exp.tex}

\section{Conclusion}
\input{ccl.tex}

\section*{Acknowledgment}
This work was supported by a grant from ANR (MACARON project
under grant number ANR-14-CE23-0003-01), by the ERC grant number 714381
(SOLARIS project), the ERC advanced grant ALLEGRO and gifts from Amazon and
Intel.

\ifCLASSOPTIONcaptionsoff
  \newpage
\fi

\begin{IEEEbiography}[{\includegraphics[width=1.1in,height=1.55in,trim=0 40 0 0,
    clip,keepaspectratio]{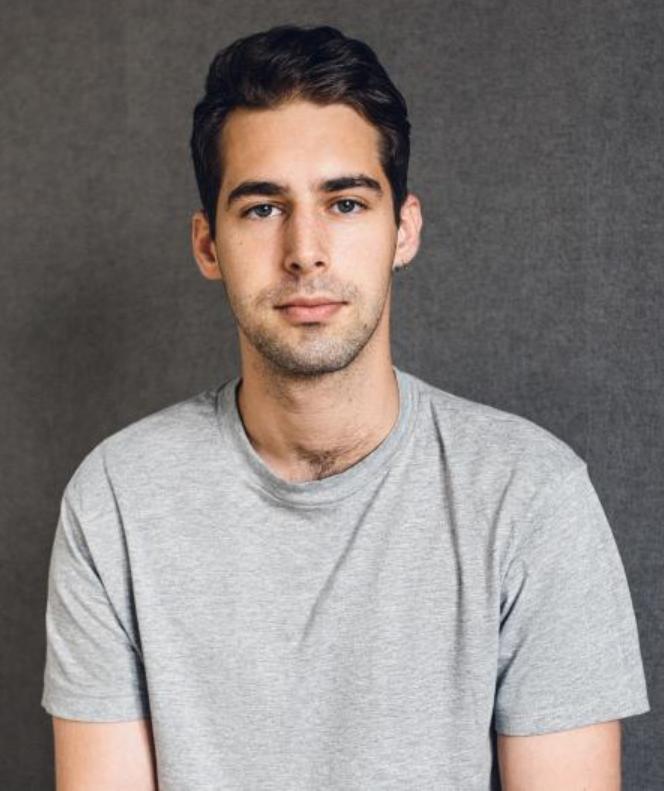}}]{Nikita Dvornik}
recieved the bachelor degree at the Moscow Institute of Physics and
Technology~(MIPT) and master degree at INP Grenoble. He is currently working
towards the PhD degree at INRIA Grenoble under supervision of Cordelia Schmid and
Julien Mairal. His research interests include scene understanding tasks, such as
object detection and semantic segmentaion, data augmentation, few-shot learning
and learning general image representations under constraints.
\end{IEEEbiography}

\begin{IEEEbiography}[{\includegraphics[width=1.1in,height=1.25in,
    clip,keepaspectratio]{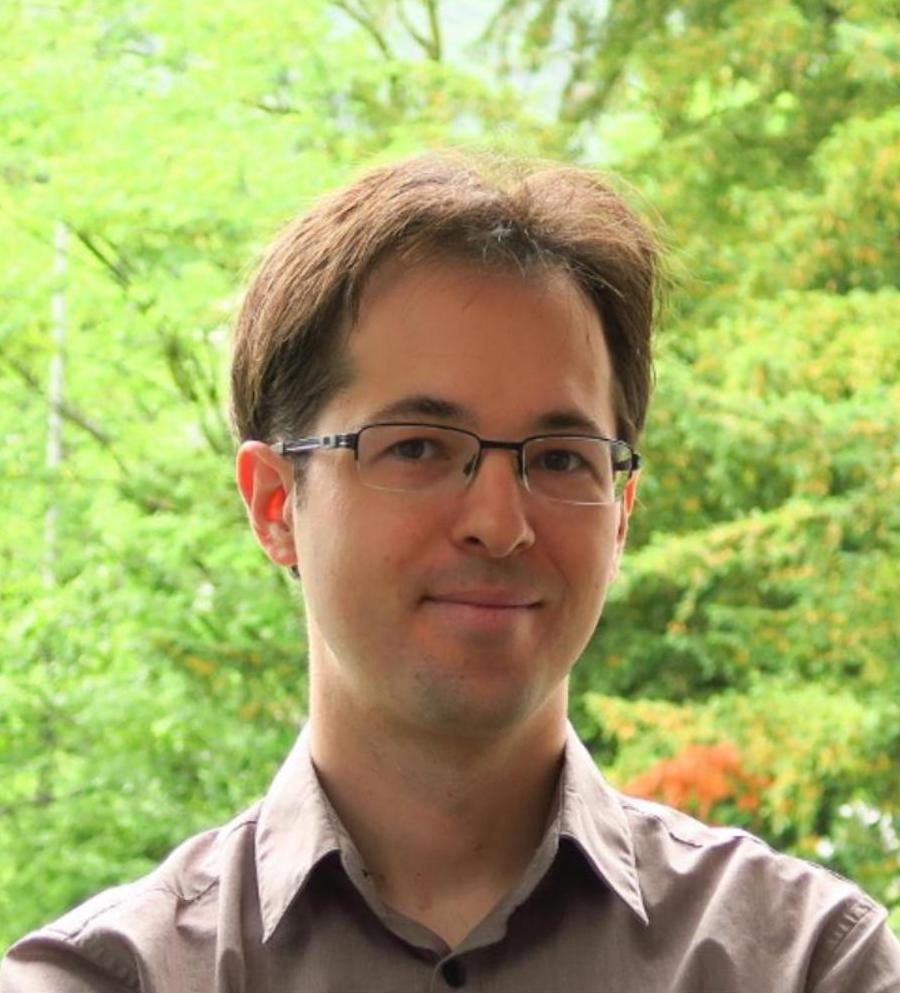}}]{Julien Mairal}
(SM’16) received the Graduate degree from the Ecole Polytechnique,
Palaiseau, France, in 2005, and the Ph.D. degree from Ecole Normale Superieure,
Cachan, France, in 2010. He was ´ a Postdoctoral Researcher at the Statistics
Department, UC Berkeley. In 2012, he joined Inria, Grenoble, France, where he is
currently a Research Scientist. His research interests include machine learning,
computer vision, mathematical optimization, and statistical image and signal
processing. In 2016, he received a Starting Grant from the European Research
Council and in 2017, he received the IEEE PAMI young research award. He was
awarded the Cor Baayen prize in 2013, the IEEE PAMI young research award in 2017
and the test-of-time award at ICML 2019.
\end{IEEEbiography}

\begin{IEEEbiography}[{\includegraphics[width=1.4in,height=1.3in,trim=0 20 0 20,
    clip,keepaspectratio]{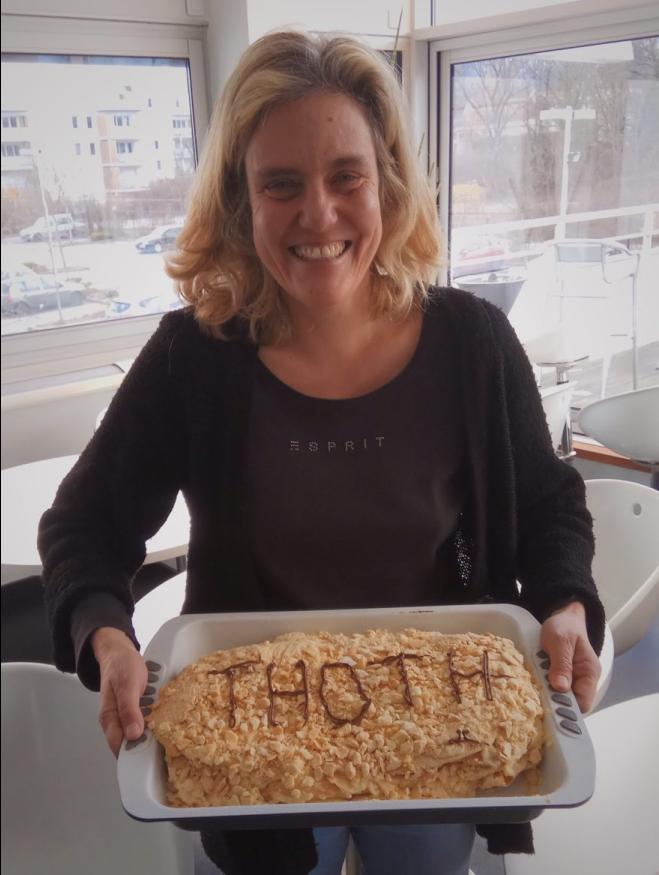}}]{Cordelia Schmid}
holds a M.S. degree in Computer Science from the University of Karlsruhe and a
Doctorate, also in Computer Science, from the Institut National Polytechnique de
Grenoble (INPG). She is a reserach director at Inria Grenoble. She has been an
editor-in-chief for IJCV (2013--2018), a program chair of IEEE CVPR 2005 and
ECCV 2012 as well as a general chair of IEEE CVPR 2015. In 2006, 2014 and 2016,
she was awarded the Longuet-Higgins prize for fundamental contributions in
computer vision that have withstood the test of time. She is a fellow of IEEE.
She was awarded an ERC advanced grant in 2013, the Humbolt research award in
2015 and the Inria \& French Academy of Science Grand Prix in 2016. She was
elected to the German National Academy of Sciences, Leopoldina, in 2017.
In 2018 she received th Koenderink prize for fundamental
contributions in computer vision that have withstood the test of time.
Starting 2018 she holds a joint appointment with Google research. 
\end{IEEEbiography}
\bibliographystyle{IEEEtran}
\bibliography{references}

\end{document}

%% file: abstract.tex
Performing data augmentation for learning deep neural networks is known to be
important for training visual recognition systems. By artificially increasing
the number of training examples, it helps reducing overfitting and improves
generalization.  While simple image transformations can already improve
predictive performance in most vision tasks, larger gains can be obtained by
leveraging task-specific prior knowledge.  In this work, we consider object
detection, semantic and instance segmentation and augment the training images by blending
objects in existing scenes, using instance segmentation annotations. We observe
that randomly pasting objects on images hurts the performance, unless the
object is placed in the right context. To resolve this issue, we propose an
explicit context model by using a convolutional neural network, which predicts
whether an image region is suitable for placing a given object or not. In our
experiments, we show that our approach is able to improve object detection,
semantic and instance segmentation on the PASCAL VOC12 and COCO datasets, with
significant gains in a limited annotation scenario, i.e. when only one category is annotated.
% when few labeled examples are available.
We also show that the method is not limited to datasets that come with expensive
pixel-wise instance annotations and can be used when only bounding boxes are
available, by employing weakly-supervised learning for instance masks
approximation.

%% file: intro.tex
\begin{figure*}[hbtp!]
\begin{center}
   \includegraphics[width=0.99\linewidth,trim=0 65 0 10,clip]{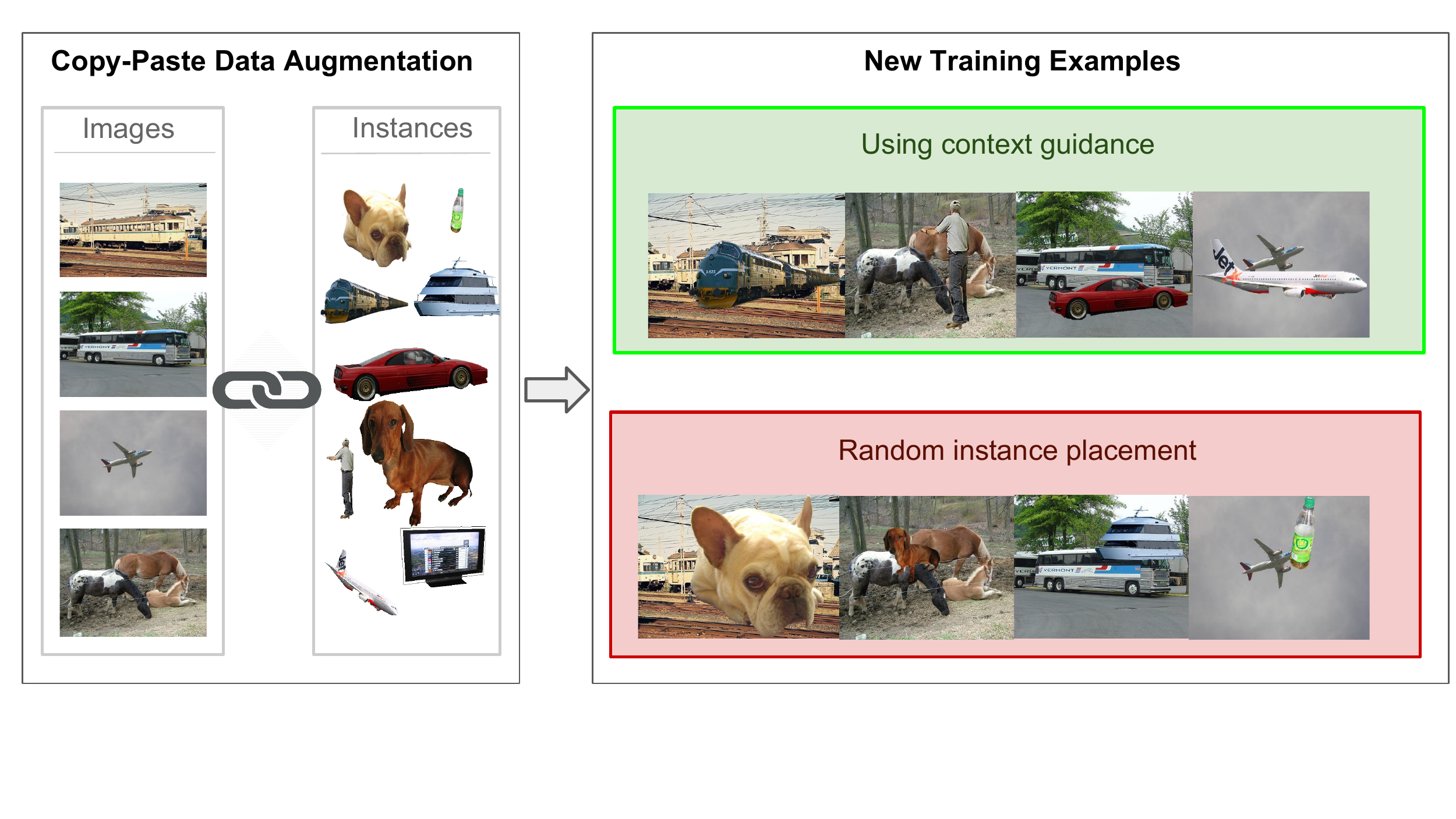}
\end{center}
\caption{\textbf{Examples of data-augmented training examples produced by our approach.}
Images and objects are taken from the VOC'12 dataset that contains segmentation annotations.
We compare the output obtained by pasting the objects with our context model vs. those obtained with random placements.
Even though the results are not perfectly photorealistic and display blending
artefacts, the visual context surrounding objects is more often correct with the explicit context model.
 }
\label{fig:segmentations2}
\end{figure*}

Convolutional neural networks (CNNs) are commonly used for scene
understanding tasks such as object detection and semantic segmentation. 
One of the major challenge to use such models is however to gather and annotate
enough training data. Various heuristics are typically used to prevent overfitting such as
DropOut~\cite{srivastava2014dropout}, penalizing the norm of the network
parameters (also called weight decay), or early stopping the optimization
algorithm. Even though the exact regularization effect of such approaches on
learning is not well understood from a theoretical point of view, these
heuristics have been found to be useful in practice.

Apart from the regularization methods related to the optimization procedure,
reducing overfitting
can be achieved with data augmentation. For most vision problems, generic input image
transformations such as cropping, rescaling, adding noise, or adjusting colors are usually helpful and
may substantially improve generalization. 
Developing more elaborate augmentation strategies requires then
prior knowledge about the task. For example, all
categories in the Pascal VOC~\cite{pascal} or ImageNet~\cite{imagenet} datasets
are invariant to horizontal flips (e.g. a flipped car is still a car). However,
flipping would be harmful for hand-written digits from the MNIST
dataset~\cite{lecun1998mnist} (e.g., a flipped ``5'' is not a digit).

A more ambitious data augmentation technique consists of leveraging segmentation
annotations, either obtained manually, or from an automatic segmentation system,
and create new images with objects placed at various positions in existing
scenes~\cite{cut_paste,gupta2016synthetic,georgakis2017synthesizing}. While not
achieving perfect photorealism, this strategy with random placements has proven
to be surprisingly effective for \emph{object instance
detection}~\cite{cut_paste}, which is a fine-grained detection task consisting
of retrieving instances of a particular object from an image collection; in
contrast, \emph{object detection} and \emph{semantic segmentation} focus on
distinguishing between object categories rather than objects themselves and have
to account for rich intra-class variability. 
For these tasks, the random-placement strategy simply does not work, as shown in
the experimental section. Placing training objects at unrealistic positions
probably forces the detector to become invariant to contextual information and
to focus instead on the object's appearance.

Along the same lines, the authors of~\cite{gupta2016synthetic} have proposed
to augment datasets for text recognition by adding text on images in a realistic
fashion. There, placing text with the right geometrical context proves to
be critical. Significant improvements in accuracy are obtained by first
estimating the geometry of the scene, before placing text on an estimated plane.
Also related, the work of \cite{georgakis2017synthesizing} is using 
successfully such a data augmentation technique for object detection in indoor scene
environments. Modeling context has been found to be critical as well and has been achieved
by also estimating plane geometry and objects are typically placed on detected
tables or counters, which often occur in indoor scenes.

In this paper, we consider more general tasks of scene understanding such as
object detection, semantic and instance segmentation, which require more generic context
modeling than estimating planes and surfaces as done for instance
in~\cite{gupta2016synthetic,georgakis2017synthesizing}. To this end, the first
contribution of our paper is methodological: we propose a context model based on
a convolutional neural network. The model estimates the likelihood of a particular object
category to be present inside a box given its neighborhood, and then
automatically finds suitable locations on images to place new objects and
perform data augmentation. A brief illustration of the output produced by this
approach is presented in Figure~\ref{fig:segmentations2}. The second
contribution is
experimental: We show with extensive tests on the COCO~\cite{coco} and VOC'12 benchmarks
using different network architectures that context modeling is in fact a key to
obtain good results for detection and segmentation tasks and that substantial
improvements over non-data-augmented baselines may be achieved when few labeled
examples are available. We also show that having expensive pixel-level
annotations of objects is not necessary for our method to work well and
demonstrate improvement in detection results when using only bounding-box
annotations to extract object masks automatically.

The present work is an extension of our preliminary work published at the
conference ECCV in 2018~\cite{dvornik2018modeling}. The main contributions
of this long version are listed below:
\begin{itemize}
  \item We show that our augmentation technique improves detection performance
  even when training on large-scale data by considering the COCO dataset for
  object detection in addition to Pascal VOC.
  \item Whereas the original data augmentation method was designed for object
    detection, we generalize it to semantic segmentation and instance segmentation.
  \item We show how to reduce the need for instance segmentation
  annotations to perform data augmentation for object detection. We employ
  weakly-supervised learning in order to automatically generate instance masks.
  \item We demonstrate the benefits of the proposed augmentation strategy for
    other object detectors than~\cite{blitznet}, by evaluating our approach with
    Faster-RCNN~\cite{faster-rcnn} and Mask-RCNN~\cite{mask_rcnn}.
%   \item We introduce modifications to the training and inference procedures of
%     the context model in order to improve its augmentation capabilities.
\end{itemize}
Our context model and the augmentation pipeline are made available as an
open-source software package (follow \url{thoth.inrialpes.fr/research/context\_aug}).

%% file: related.tex
In this section, we discuss related work for visual context modeling, data
augmentation for object detection and semantic segmentation and methods suitable
for automatic object segmentation.

\textbf{Modeling visual context for object detection.}
Relatively early, visual context has been modeled by computing statistical
correlation between low-level features of the global scene and descriptors
representing an object~\cite{torralba2001statistical,torralba2003contextual}.
Later, the authors of~\cite{felzenszwalb2010object} introduced a simple context
re-scoring approach operating on appearance-based detections. To encode more
structure, graphical models were then widely used in order to jointly model
appearance, geometry, and contextual
relations~\cite{choi2010exploiting,gould2009decomposing}. Then, deep learning
approaches such as convolutional neural networks started to be
used~\cite{faster-rcnn,fast-rcnn,ssd}; as mentioned previously, their features
already contain implicitly contextual information. Yet, the work
of~\cite{chu2018deep} explicitly incorporates higher-level context clues and
combines a conditional random field model with detections obtained by
Faster-RCNN. With a similar goal, recurrent neural networks are used
in~\cite{bell2015inside} to model spatial locations of discovered objects.
Another complementary direction in context modeling with convolutional neural
networks use a deconvolution pipeline that increases the field of view of
neurons and fuse features at different
scales~\cite{bell2015inside,blitznet,dssd}, showing better performance
essentially on small objects. The works
of~\cite{divvala2009empirical,barnea2017utility} analyze different types of
contextual relationships, identifying the most useful ones for detection, as
well as various ways to leverage them. However, despite these efforts, an
improvement due to purely contextual information has always been relatively
modest~\cite{yu2016role,yao2010modeling}.

\textbf{Modeling visual context for semantic segmentation.} While object
detection operates on image's rectangular regions, in semantic segmentation the
neighboring pixels with similar values are usually organized together in
so-called superpixels~\cite{ren2003learning}. This allows defining contextual
relations between such regions. The work of~\cite{he2006learning}
introduces ``context clusters'' that are discovered and learned from region
features. They are later used to define a specific class model for each context
cluster. In the work of~\cite{yang2014context} the authors tile an image with
superpixels at different scales and use this representation to build global and
local context descriptors. The work of~\cite{shotton2006textonboost} computes
texton features~\cite{leung2001representing} for each pixel of an image and
defines shape filers on them. This enables the authors to compute local and
middle-range concurrence statistics and enrich region features with context
information. Modern CNN-based methods on the contrary rarely define an explicit
context model and mostly rely on large receptive fields~\cite{long2015fully}.
Moreover, by engineering the network's architecture one can explicitly require
local pixel descriptors used for classification to carry global image
information too, which enables reasoning with context. To achieve this goal
encoder-decoder architectures~\cite{badrinarayanan2015segnet,blitznet} use
deconvolutional operations to propagate coarse semantic image-level information
to the final layers while refining details with local information from earlier
layers using skip-connections. As an alternative, one can use dilated
convolutions~\cite{yu2015multi,chen2018deeplab} that do not down-sample the
representation but rather up-sample the filters by introducing ``wholes'' in
them. Doing so is computationally efficient and allows to account for global image
statistics in pixel classification. Even though visual context is implicitly
present in the networks outputs, it is possible to define an explicit context
model~\cite{chen2018deeplab,lin2018exploring} on top of them. This usually results
in moderate improvement in model's accuracy.

\textbf{Data augmentation for object detection and segmentation.}
Data augmentation is a major tool to train deep neural networks. It varies from
trivial geometrical transformations such as horizontal flipping, cropping with
color perturbations, and adding noise to an image~\cite{krizhevsky2012imagenet},
to synthesizing new training images~\cite{frid2018synthetic,peng2015learning}.
Some recent object detectors~\cite{ssd,yolo,blitznet} benefit from standard data
augmentation techniques more than others~\cite{faster-rcnn,fast-rcnn}. The
performance of Fast- and Faster-RCNN could be for instance boosted by simply
corrupting random parts of an image in order to mimic
occlusions~\cite{random_erase}. The field of semantic segmentation is enjoying a
different trend---augmenting a dataset with synthetic images. They could be
generated using extra annotations~\cite{sakaridis2018semantic}, come from a
purely synthetic dataset with dense
annotations~\cite{handa2016understanding,mccormac2017scenenet} or a simulated
environment~\cite{qiu2016unrealcv}. For object detection, recent works such
as~\cite{karsch2011rendering,movshovitz2016useful,su2015render} also build and
train their models on purely synthetic rendered 2d and 3d scenes. However, a
major difficulty for models trained on synthetic images is to guarantee that
they will generalize well to real data since the synthesis process introduces
significant changes of image statistics \cite{peng2015learning}. This problem
could be alleviated by using transfer-learning techniques such as
\cite{sankaranarayanan2018learning} or by improving photo-realism of synthetic
data \cite{barth2018improved,sixt2018rendergan}. To address the same issue, the
authors of \cite{gupta2016synthetic} adopt a different direction by pasting real
segmented object into natural images, which reduces the presence of rendering
artefacts. For object instance detection, the
work~\cite{georgakis2017synthesizing} estimates scene geometry and spatial
layout, before synthetically placing objects in the image to create realistic
training examples. In \cite{cut_paste}, the authors propose an even simpler
solution to the same problem by pasting images in random positions but modeling
well occluded and truncated objects, and making the training step robust to
boundary artifacts at pasted locations. In contrast to this method,
our approach does not choose pasted locations at random but uses
an explicit context model. We found this crucial to improve general object
detection.

\begin{figure*}[t!]
\begin{center}
  \includegraphics[width=1.09\linewidth,trim=32 50 0 50,clip]{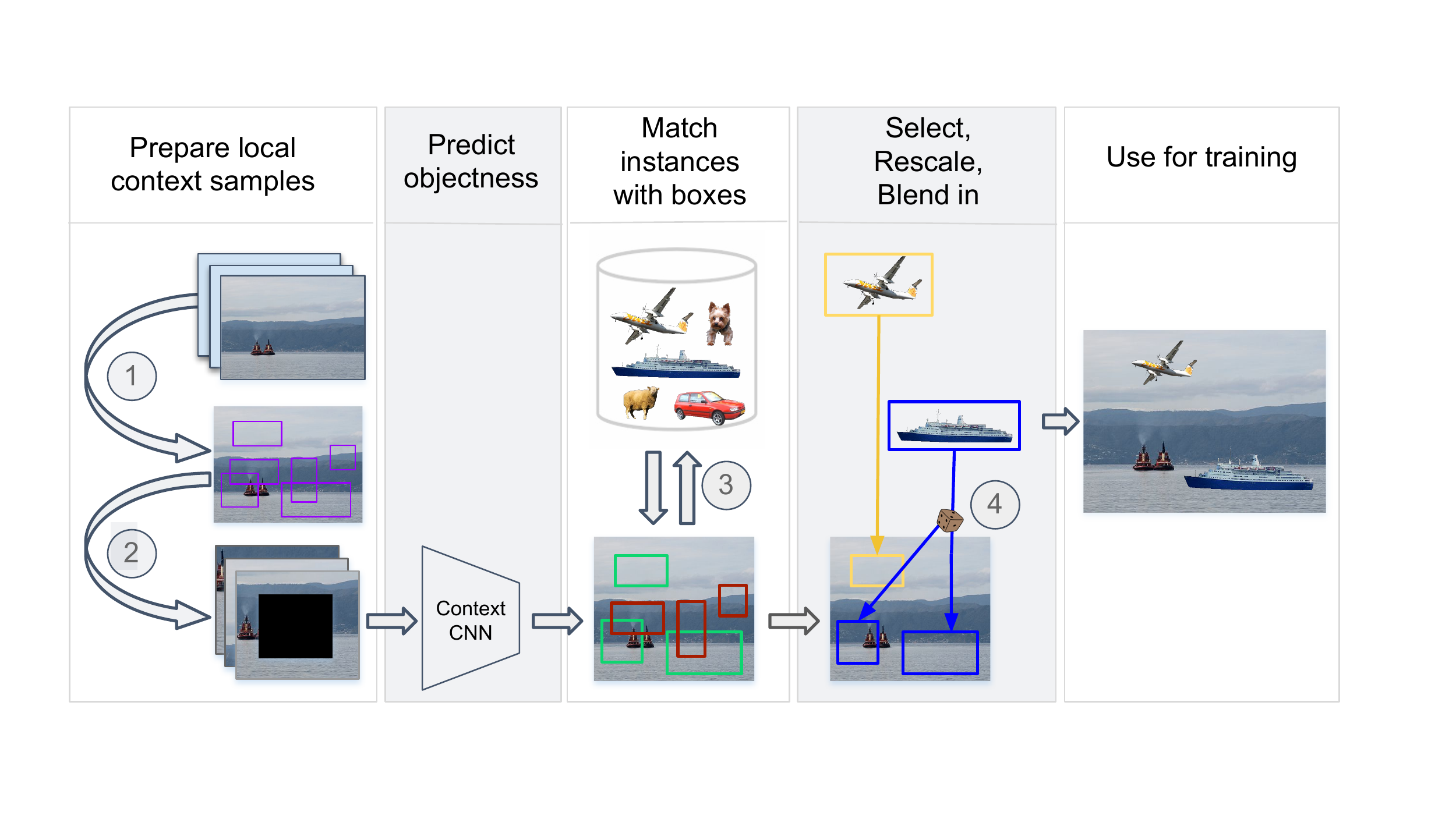}
\end{center}
\vspace{-0.5cm}
\caption{\textbf{Illustration of our data augmentation approach.}
  We select an image for augmentation and 1) generate 200 candidate boxes that cover
  the image. Then, 2) for each box we find a neighborhood that contains the box
  entirely, crop this neighborhood and mask all pixels falling inside
  the bounding box; this ``neighborhood'' with masked pixels is then fed to the
  context neural network module and 3) object instances are matched to boxes that
  have high confidence scores for the presence of an object category. 4) We select
  at most two instances that are rescaled and blended into the selected bounding
  boxes. The resulting image is then used for training the object detector.}
\label{fig:diagram}
\end{figure*}

\textbf{Automatic Instance Segmentation}
The task of instance segmentation is challenging and requires considerable
amount of annotated data~\cite{coco} in order to achieve good results.
Segmentation annotations are the most labor-demanding since they require
pixel-level precision. The need to distinguish between instances of one class
makes annotating ``crowd scenes'' extremely time-consuming. If data for this
problem comes without labels, tedious and expensive process of annotation may
suggests considering other solutions that do not require full supervision. The
work of
\cite{liao2012building} uses various image statistics and hand-crafted
descriptors that do not require learning along with annotated image tags, in order
to build a segmentation proposal system. With very little supervision, they learn
to descriminate between ``good'' and ``bad'' instance masks and as a
result are able to automatically discover good quality instance segments within the
dataset. As an alternative, one can use weakly-supervised methods to estimate
instance masks. The authors of \cite{zhou2018weakly} use only category
image-level annotations in order to train an object segmentation system. This is
done by exploiting class-peak responses obtained using pre-trained
classification network and propagating them spatially to cover meaningful
image segments. It is beneficial to use instance-level
annotations, such as object boxes and corresponding categories, if those
are available, in order to improve the system's performance. The work
of \cite{khoreva2017simple} proposes a rather simple yet efficient framework for
doing so. By providing the network with extra information, which is a
rectangular region containing an object, a system learns to discover instance
masks automatically inside those regions. Alternatively, the system could be
trained to provide semantic segmentation masks in a weakly-supervised fashion.
Together with bounding boxes, one may use it to approximate instance masks.

%% file: approach.tex
In this section, we present a simple experiment to motivate our context-driven
data augmentation, and present the full pipeline in details. We start by
describing a naive solution to augmenting an object detection dataset, which is
to perform copy-paste data augmentation agnostic to context by placing objects
at random locations. Next, we explain why it fails for our task and propose a
natural solution based on explicit context modeling by a CNN. We show how to
apply the context model to perform augmentation for detection and segmentation
tasks and how to blend the object into existing scenes. The full pipeline is
depicted in Figure.~\ref{fig:diagram}.

\subsection{Copy-paste Data Augmentation with Random Placement is not Effective
  for Object Detection}\label{sec:random_copy-paste}
In \cite{cut_paste}, data augmentation is performed by positioning segmented objects 
at random locations in new scenes. As mentioned previously, the strategy
was shown to be effective for object \emph{instance} detection, as soon as an appropriate
procedure is used for preventing the object detector to overfit blending
artefacts---that is, the main difficulty is to prevent the detector to ``detect
artefacts'' instead of detecting objects of interest.
This is achieved by using various blending strategies to smooth object boundaries such as Poisson
blending~\cite{perez2003poisson}, and by adding ``distractors'' -
objects that do not belong to any of the dataset categories, but which are also
synthetically pasted on random backgrounds. With distractors, artefacts occur
both in positive and negative examples, for each of the categories, preventing
the network to overfit them.
According to~\cite{cut_paste}, this strategy can bring substantial improvements
for the object instance detection/retrieval task, where modeling the fine-grain
appearance of an object instance seems to be more important than modeling
visual context as in the general category object detection task.

Unfortunately, the augmentation strategy described above does not improve the
results on the general object detection task and may even hurt the performance
as we show in the experimental section. To justify the initial claim, we
follow~\cite{cut_paste} as close as possible and conduct the following
experiment on the PASCAL VOC12 dataset~\cite{pascal}. Using provided instance
segmentation masks we extract objects from images and store them in a so-called
instance-database. They are used to augment existing images in the training
dataset by placing the instances at random locations. In order to reduce
blending artifacts we use one of the following strategies: smoothing the edges
using Gaussian or linear blur, applying Poisson blending~\cite{perez2003poisson}
in the segmented region, blurring the whole image by simulating a slight camera
motion or leaving the pasted object untouched. As distractors, we used objects
from the COCO dataset~\cite{coco} belonging to categories not present
in PASCAL VOC~\footnote{Note that external data from COCO was used only in
this preliminary experiment and not in the experiments reported later in
Section~\ref{sec:exp}.}.

\begin{figure}[btp!]
\begin{center}
  \includegraphics[width=0.99\linewidth,trim=90 115 230 30,clip]{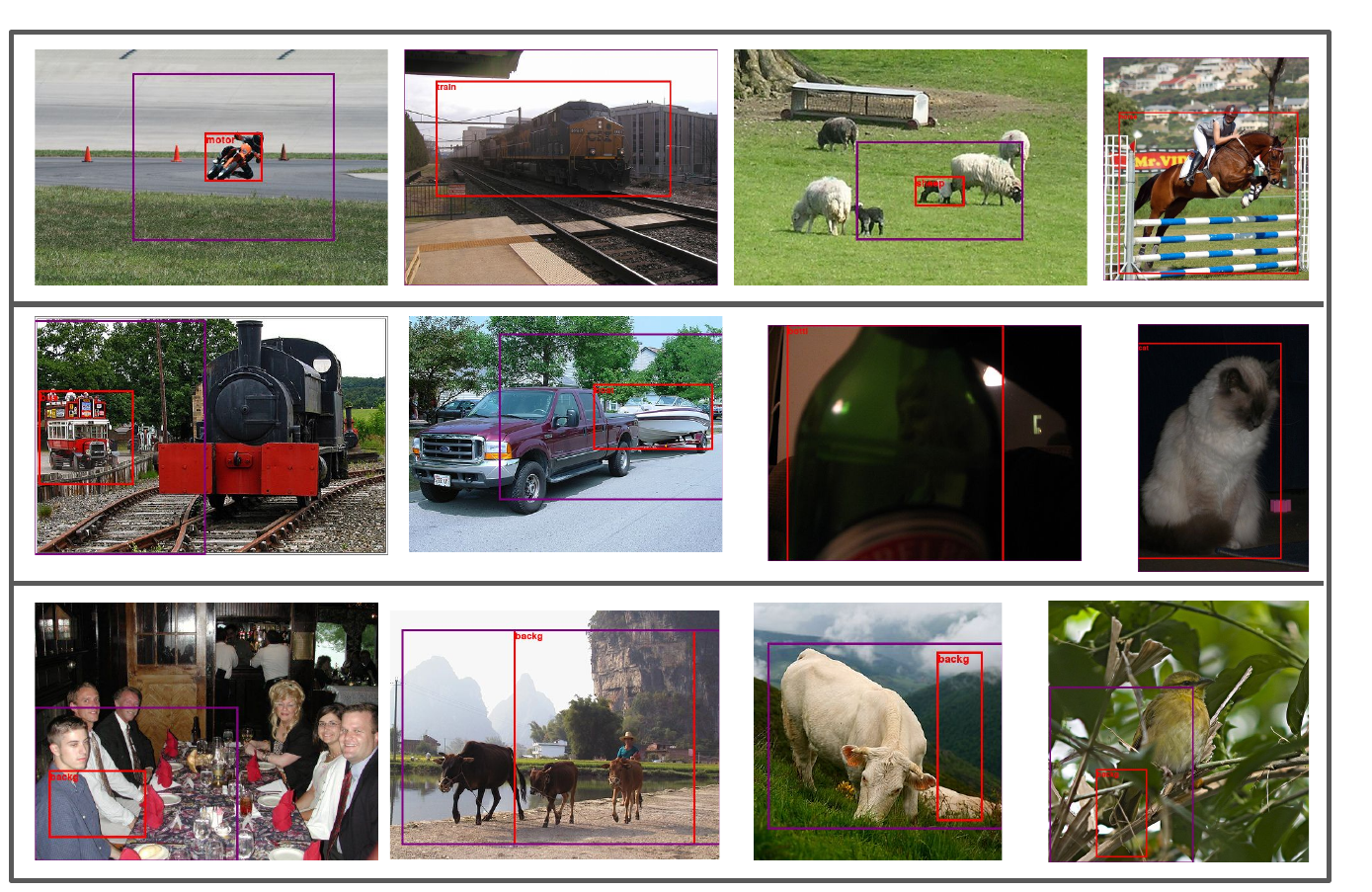}
\end{center}
\caption{\textbf{Contextual images - examples of inputs to the context model}.
  A subimage bounded by a magenta box is used as an input to the context model
  after masking-out the object information inside a red box.
  The top row lists examples of positive samples encoding 
  real objects surrounded by regular and predictable context.
  Positive training examples with ambiguous or uninformative context are
  given in the second row.
  The bottom row depicts negative examples enclosing background.
  This figure shows that contextual images could be ambiguous to classify
  correctly and the task of predicting the category given only the context is
  challenging.}
\label{fig:context_input}
\end{figure}

For any combination of blending strategy, by using distractors or not, the naive
data augmentation approach with random placement did not improve upon the
baseline without data augmentation for the classical object detection task. A
possible explanation may be that for instance object detection, the detector
does not need to learn intra-class variability of object/scene representations
and seems to concentrate only on appearance modeling of specific instances,
which is not the case for category-level object detection. This experiment was
the key motivation for proposing a context model, which we now present.

\subsection{Explicit Context Modeling by CNN}
The core idea behind the proposed method is that it is possible to some extent
to guess the category of an object just by looking at its visual surroundings.
That is precisely what we are modeling by a convolutional neural network, which
takes contextual neighborhood of an object as input and is trained to predict
the object's class. Here, we describe the training data and the learning
procedure in more details.

\begin{figure}[t!]
\begin{center}
  \includegraphics[width=1.09\linewidth,trim=110 150 65 10,clip]{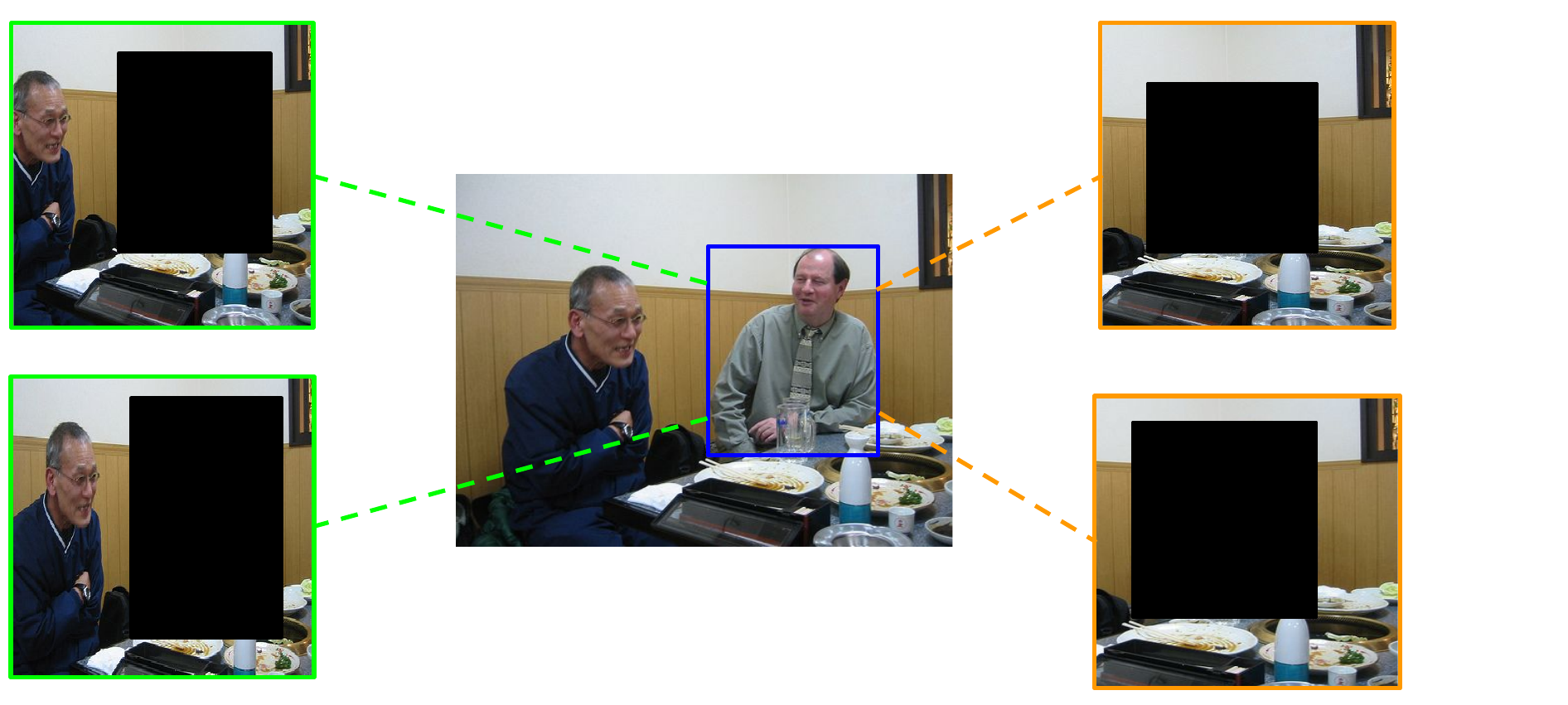}
\end{center}
\vspace{-0.5cm}
\caption{\textbf{Different contextual images obtained from a single bounding box.}
  A single ground-truth bounding box (in blue) is able to generate a set of
  different context images (in green and orange) by varying the size of the
  initial box and the context neighborhood. While the orange contextual images
  may be recognized as a chair, the green ones make it more clear that the person
  was masked out. This motivates the need to evaluate several context images for
  one box during the context estimation phase.}
\label{fig:contextual_image_variations}
\end{figure}

\textit{Contextual data generation.}\label{sec:context_inputs}
In order to train the contextual model we use a dataset that comes with
bounding box and object class annotations. Each ground-truth bounding box in the
dataset is able to generate positive ``contextual images'' that are used as
input to the system. As depicted in the Figure~\ref{fig:context_input}, a
``contextual image'' is a sub-image of an original training image, fully enclosing the
selected bounding box, whose content is masked out. Such a contextual image only carries
information about visual neighborhood that defines middle-range context and no
explicit information about the deleted object.
% In order to increase the amount of training samples, we generate multiple
% context images from one corresponding bounding box by randomly varying the size
% of the context neighborhood and up-scaling the box to be cut out,
One box is able to generate multiple different context images,
as illustrated in Figure~\ref{fig:contextual_image_variations}. Background
``contextual images'' are generated from bounding boxes that do not contain an
object
and are formally defined in~\cite{dvornik2018modeling}.
% More formally, we build contextual images from bounding boxes whose maximum
% intersection over union with any of the object boxes in an image is smaller than
% 0.3.
To prevent distinguishing between positive and background
images only by looking at the box shape and to force true visual context
modeling, we estimate the shape distribution of positive boxes and sample the
background ones from it.
Precisely, we estimate the joint distribution of scale $s$ and aspect ratio $a$
with a two-dimensional histogram, as described in~\cite{dvornik2018modeling},
% The shape is fully characterized by scale $s$ and
% aspect ratio $a$. We model their joint distribution empirically by building a 2d
% histogram $30 \times 30$, smoothing it linearly between the bins
and we draw a pair $(s, a)$ from this distribution in order to construct a
background box.
Since in natural images there is more background boxes than the ones actually
containing an object, we address the imbalance by sampling more background
boxes, following sampling strategies in \cite{dvornik2018modeling,faster-rcnn}.

\textit{Model training.}
Given the set of all contexts, gathered from all training data, we train a
convolutional neural network to predict the presence of each object in the
masked bounding box. The input to the network are the ``contextual images''
obtained during the data generation step. These contextual images are resized to
$300 \times 300$ pixels, and the output of the network is a label in
$\{0,1,...,C\}$, where $C$ is the number of object categories. The
$0$-th class represents background and corresponds to a negative ``context
image''. For such a multi-class image classification problem, we use the
classical ResNet50 network~\cite{resnet} pre-trained on ImageNet, and change the
last layer to be a softmax with $C+1$ activations (see experimental section for
details).

\begin{figure*}[btp!]
\begin{center}
  \includegraphics[width=0.99\linewidth,trim=10 0 2 4,clip]{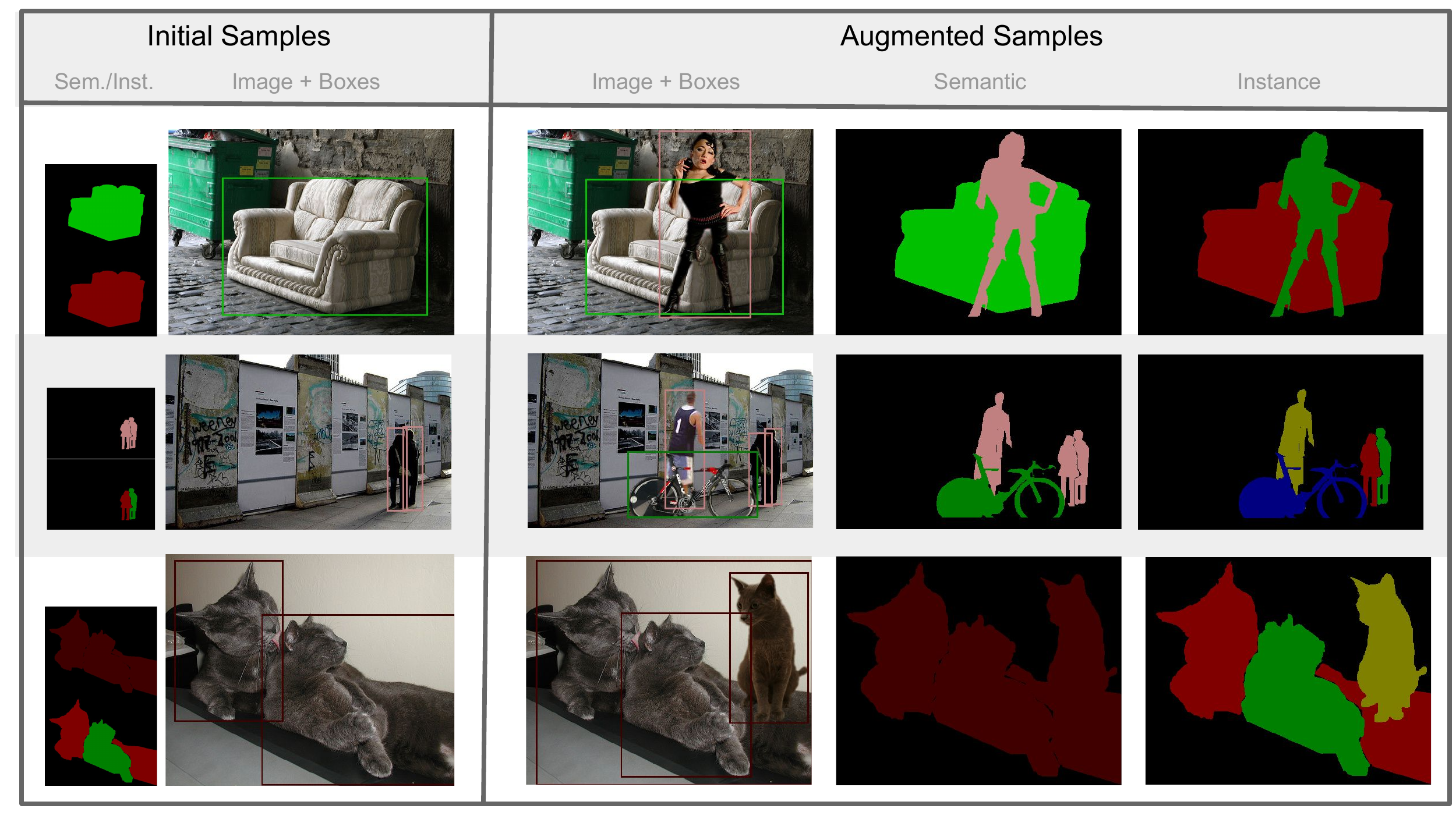}
\end{center}
\caption{\textbf{Data augmentation for different types of annotations.}
  The first column contains samples from the training dataset with corresponding
  semantic/instance segmentation and bounding box annotations. Columns 2-4
  present the result of applying context-driven augmentation to the initial
  sample with corresponding annotations.}
\label{fig:augmented_samples}
\end{figure*}

\subsection{Context-driven Data Augmentation}\label{subsec:context_da}
Once the context model is trained, we use it to provide locations where to
paste objects. In this section, we elaborate on the context network inference
and describe the precise procedure used for blending new objects into existing
scenes.

\textit{Selection of candidate locations for object placement.}
A location for pasting an object is represented as a bounding box. For a single
image, we sample 200 boxes at random from the shape distribution used in
\ref{sec:context_inputs} and later select the successful placement candidates among
them. These boxes are used to build corresponding contextual images, that we
feed to the context model as input. As output, the model provides a set of
scores in range between 0 and 1, representing the presence likelihood of each
object category in a given bounding box, by considering its visual surrounding.
The top scoring boxes are added to the final candidate set.
Since the model takes into account not only the visual surroundings but a box's
geometry too, we need to consider all possible boxes inside
an image to maximize the recall. However this is too costly and using 200
candidates was found to provide good enough bounding boxes among the top scoring
ones.\\
After analyzing the context model's output we made the following observation: if an
object of category $c$ is present in an image it is a confident signal for
the model to place another object of this class nearby. The model ignores this
signal only if no box of appropriate shape was sampled in the object's
neighborhood. This often happens when only 200 candidate locations are sampled;
however, evaluating more locations would introduce a computational overhead. To
fix this issue, we propose a simple heuristic, which consists of drawing boxes
in the neighborhood of this object and adding them to the final candidate set.
The added boxes have the same geometry (up to slight distortions) as the
neighboring object's box.

\textit{Candidate scoring process.}
As noted before, we use the context model to score the boxes by using its
softmax output. Since the process of generating a contextual image is not
deterministic, predictions on two contextual images corresponding to the same
box may differ substantially, as illustrated in
Figure~\ref{fig:contextual_image_variations}. We alleviate this effect by
sampling 3 contextual images for one location and average the predicted scores.
After the estimation stage we retain the boxes where an object category has
score greater than $0.7$; These boxes together with the candidates added at the
previous step form the final candidate set that will be used for object
placement.

\textit{Blending objects in their environment.}
Whenever a bounding box is selected by the previous procedure, we need to blend
an object at the corresponding location. This step follows closely the findings
of~\cite{cut_paste}. We consider different types of blending techniques
(Gaussian or linear blur, simple copy-pasting with no post-processing, or
generating blur on the whole image to imitate motion), and randomly choose one
of them in order to introduce a larger diversity of blending artefacts.
Figure~\ref{fig:blendings} presents the blending techniques mentioned above.
We also do not consider Poisson blending in our approach, which was considerably
slowing down the data generation procedure. Unlike~\cite{cut_paste} and unlike
our preliminary experiment described in Section~\ref{sec:random_copy-paste}, we
do not use distractors, which were found to be less important for our task than
in~\cite{cut_paste}. As a consequence, we do not need to exploit external data
to perform data augmentation.

\textit{Updating image annotation.}
Once an image is augmented by blending in a new object, we need to modify the annotation accordingly. In this work, we consider data augmention for both object detection and
semantic segmentation,  as illustrated in Figure~\ref{fig:augmented_samples}. Once a new object is
placed in the scene, we generate a bounding box for object detection
 by drawing the tightest box around that object. In
case where an initial object is too occluded by the blended one, i.e.
the IoU between their boxes is higher than 0.8, we delete the bounding box of
the original object from the annotations. For semantic segmentation, we start by
considering augmentation on instance masks (Figure~\ref{fig:augmented_samples},
column 4) and then convert them to semantic masks (Figure~\ref{fig:augmented_samples},
column 3). If a new instance occludes more than $80\%$ of an object already
present in the scene, we discard annotations for all pixels belonging to
the latter instance. To obtain semantic segmentation masks from instance
segmentations, each instance pixel is labeled with the corresponding objects
class.

%% file: exp.tex
In this section, we use the proposed context model to augment object detection
and segmentation datasets. We start by presenting
experimental and implementation details in Sections~\ref{sec:datasets}
and \ref{sec:details} respectively. In Section \ref{sec:prelim} we present a
preliminary experiment that motivates the proposed solution. In
Sections~\ref{sec:single}~and~\ref{sec:multiple} we study the effect of
context-driven data augmentation when augmenting an object
detection dataset. For this purpose we consider the Pascal VOC12 dataset that has
instance segmentation annotations and  we demonstrate the applicability of
our method to different families of object detectors. We study the
scalability of our approach in Section~\ref{exp:coco} by using the
COCO dataset for object detection and instance segmentation. We show benefits of
our method in Section~\ref{sec:segmentation} by augmenting the VOC12 for
semantic segmentation. In Section \ref{exp:weak}, we use
weakly-supervised learning for estimating object masks and evaluate our approach
on the Pascal VOC12 dataset using only bounding box annotations.
Finally, Section~\ref{exp:context_quality} studies how the amount of data
available for training the context model influences the final detection
performance.

\begin{figure}[btp!]
\begin{center}
  \includegraphics[width=0.99\linewidth,trim=100 0 250 320,clip]{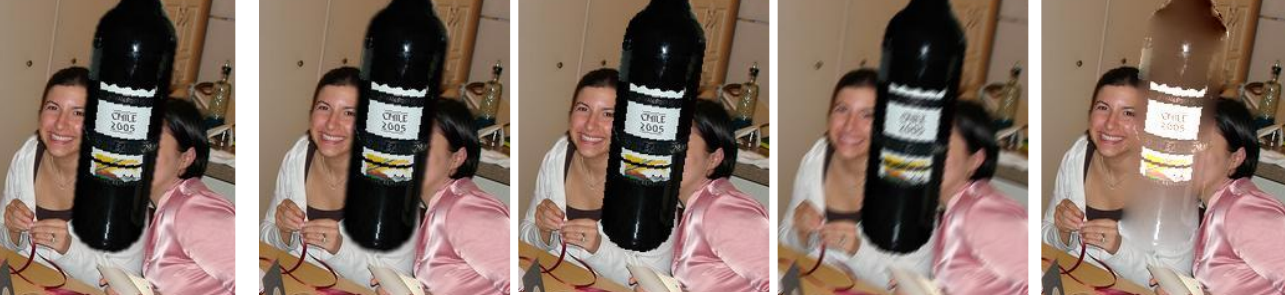}
\end{center}
\caption{\textbf{Different kinds of blending used in experiments.}
  From left to right: linear smoothing of boundaries, Gaussian smoothing, no
  processing, motion blur of the whole image, Poisson blending \cite{perez2003poisson}.}
\label{fig:blendings}
\end{figure}

\subsection{Dataset, Tools, and Metrics} \label{sec:datasets}

\textit{Datasets.}
In our experiments, we use the Pascal VOC'12~\cite{pascal} and COCO~\cite{coco}
datasets. In the VOC'12 dataset, we only consider a subset that contains
segmentation annotations. The training set contains $1\,464$ images and is
dubbed \texttt{VOC12train-seg} later in the paper. Following standard practice,
we use the test set of VOC'07 to evaluate the detection performance, which
contains $4\,952$ images with the same 20 object categories as VOC'12. We call
this image set \texttt{VOC07-test}. When evaluating segmentation performance, we
use the validation set of the VOC'12 annotated with segmentation masks
\texttt{VOC12val-seg} that contains $1\,449$ images.\\ The COCO dataset
\cite{coco} is used for large-scale object detection experiments. It includes 80
object categories for detection and instance segmentation. For both tasks, there
are 118K images for training that we denote as \texttt{COCO-train2017} and 5K
for validation and testing denoted as \texttt{COCO-val2017}.

\textit{Models.}
To test our data-augmentation strategy we chose a single model capable of
performing both object detection and semantic segmentation. BlitzNet
\cite{blitznet} is an encoder-decoder architecture,
which is able to solve either of the tasks, or both simultaneously if trained
with box and segmentation annotations together. The open-source
implementation is available online. If used to solve the detection task,
BlitzNet achieves close to the state-of-the-art results ($79.1\%$ mAP) on
\texttt{VOC07-test} when trained on the union of the full training and
validation parts of VOC'07 and VOC'12, namely \texttt{VOC07-train+val} and
\texttt{VOC12train+val} (see~\cite{blitznet}); this network is similar to the
DSSD detector of \cite{dssd} that was also used in the Focal Loss paper
\cite{focal_loss}. When used as a segmentor, BlitzNet resembles the classical
U-Net architecture~\cite{ronneberger2015u} and also achieves results comparable
to the state-of-the-art on VOC'12-test set ($75.5\%$ mIoU). The advantage of
such class of models is that it is relatively fast (it may work in real time)
and supports training with big batches of images without further modification.\\
To make the evaluation extensive, we also consider a different region-based
class of detectors. For that purpose we employ an open-source implementation of
Faster-RCNN \cite{jjfaster2rcnn} which uses ResNet50~\cite{resnet}
architecture as a feature extractor.
Finally, when tackling object detection and instance segmentation on COCO,
we use Mask-RCNN~\cite{mask_rcnn} that solves both tasks simultaneously. For
each region proposal the network outputs estimated class probabilities, regressed
box offsets and a predicted instance mask. We run the official implementation
of~\cite{Detectron2018} that uses ResNet50 as a backbone, followed by an
FPN~\cite{lin2016feature} module. This setup corresponds to the current state of the art
in object detection and instance segmentation.

\textit{Evaluation metric.}
In VOC'07, a bounding box is considered to be correct if its Intersection over
Union (IoU) with a ground truth box is higher than 0.5. The metric for
evaluating the quality of object detection and instance segmentation for one
object class is the average precision (AP). Mean Average Precision (mAP) is
used to report the overall performance on the dataset. Mean Intersection Over
Union (mIoU) is used to measure performance on semantic segmentation.

\subsection{Implementation Details}\label{sec:details}

\begin{figure*}[hbtp!]
\begin{center}
  \includegraphics[width=0.99\linewidth,trim=20 70 30 0,clip]{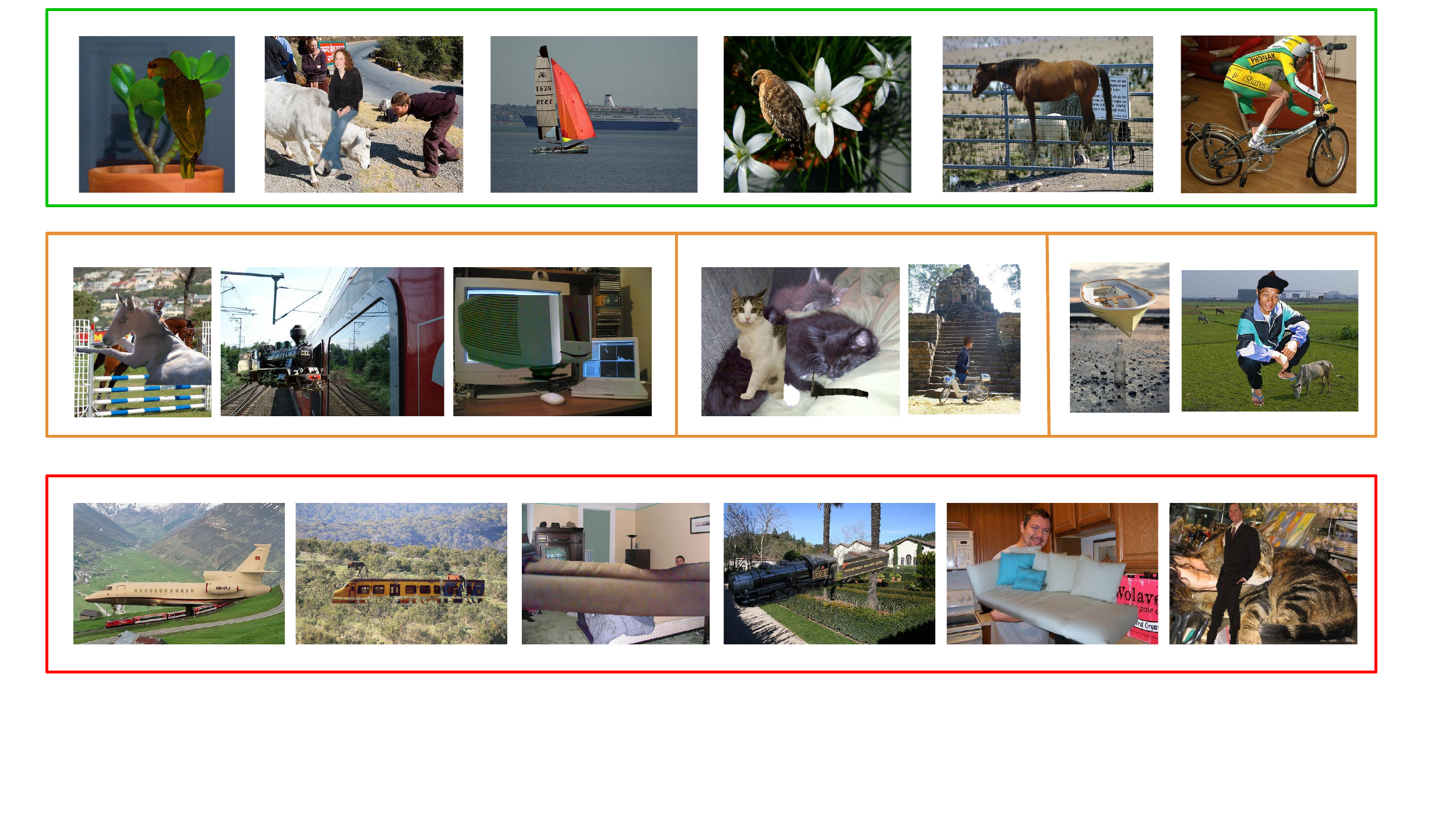}
\end{center}
% \vspace*{-0.4cm}
\caption{\textbf{Examples of instance placement with context model guidance.}
  The figure presents samples obtained by placing a matched examples into the
   box predicted by the context model. The top row shows generated images that are
   visually almost indistinguishable from the real ones. The middle row presents
   samples of good quality although with some visual artifacts. For the two
   leftmost examples, the context module proposed an appropriate object class, but
   the pasted instances do not look visually appealing. Sometimes, the scene does
   not look natural because of the segmentation artifacts as in the two middle
  images. The two rightmost examples show examples where the category seems to be
  in the right environment, but not perfectly placed.
  The bottom row presents some failure cases.}
\label{fig:context_output}
\end{figure*}

\textit{Training the context model.} After preparing the ``contextual images'' as
described in \ref{sec:context_inputs}, we re-scale them to the standard size
$300 \times 300$ and stack them in batches of size 32. We use ResNet50~\cite{resnet} with
ImageNet initialization to train a contextual model in all our experiments.
Since we have access only to the training set at any stage of the pipeline we
define two strategies for training the context model. When the amount of
positive samples is scarce, we train and apply the model on the same data. To
prevent overfitting, we use early stopping. In order to determine when to stop
the training procedure, we monitor both training error on our training set and
validation error on the validation set. The moment when the loss curves start
diverging noticeably is used as a stopping point. We call this training setting
``small-data regime''. When the size of the training
set is moderate and we are in ``normal-data regime'', we split it in two parts
ensuring that for each class, there is a similar number of positive examples in
both splits. The context model is trained on one split and applied to
another one. We train the model with ADAM optimizer \cite{adam} starting with
learning rate $10^{-4}$ and decreasing it by the factor of 10 once during the
learning phase. The number of steps depends on a dataset.
We sample 3 times more background contextual images, as noted in
Section~\ref{sec:context_inputs}. Visual examples of augmented images produced
when using the context model are presented in Figure~\ref{fig:context_output}.
Overall, training the context model is about 4-5 times faster than training the
detector.

\textit{Training detection and segmentation models.}
In this work, the BlitzNet model takes images of size $300 \times 300$ as an
input and produces a task-specific output. When used as a detector, the output
is a set of candidate object boxes with classification scores and in case of
segmentation it is an estimated semantic map of size $75\times75$; like our
context model, it uses ResNet50 \cite{resnet} pre-trained on ImageNet as a
backbone. The models are trained by following~\cite{blitznet}, with the ADAM
optimizer \cite{adam} starting from learning rate $10^{-4}$ and decreasing it
later during training by a factor 10 (see
Sections~\ref{sec:det_voc}~and~\ref{sec:segmentation} for number of epochs used
in each experiment). In addition to our data augmentation approach obtained by
copy-pasting objects, all experiments also include classical data augmentation
steps obtained by random-cropping, flips, and color transformations,
following~\cite{blitznet}. For the Faster-RCNN detector training, we consider
the classical model of \cite{faster-rcnn} with ResNet50 backbone and closely
follow the instructions of~\cite{jjfaster2rcnn}. On the Pascal VOC12 dataset,
training images are rescaled to have both sides between 600 and 1000 pixels
before being passed to the network. The model is trained with the Momentum
optimizer for 9 epochs in total. The starting learning rate is set to $10^{-2}$
and divided by 10 after first 8 epochs of training. When using
Mask-RCNN~\cite{mask_rcnn}, the images are rescaled to have a maximum size of
1333 pixel on one side or a minimum one of 800 pixels. \red{Following the
  original implementation~\cite{Detectron2018}, training and evaluation
is performed on 2 GPUs, where images are grouped in batches of size 16}. We
set the starting learning rate to $2 \cdot 10^{-2}$ which is decreased by a
factor of 10 twice later during training. For both Faster-RCNN and Mask-RCNN
standard data augmentation includes only horizontal flipping.

\textit{Selecting and blending objects.}
Since we widely use object instances extracted from the training images in all
our experiments, we create a database of objects cut out from
the \texttt{VOC12train-seg} or
\texttt{COCO-train} sets to quickly access them during training. For a
given candidate
box, an instance is considered as matching if after scaling it by a factor in
$[0.5, 1.5]$ the re-scaled instance's bounding box fits inside the candidate's
one and takes at least 80\% of its area. The scaling factor is kept close to 1
not to introduce scaling artefacts. When blending the objects into the new
background, we follow~\cite{cut_paste} and use randomly one of the following
methods: adding Gaussian or linear blur on the object boundaries, generating
blur on the whole image by imitating motion, or just paste an image with no
blending. By introducing new instances in a scene we may also introduce heavy
occlusions of existing objects. The strategy for resolving this issue depends on
the task and is clarified in Sections \ref{sec:det_voc} and \ref{sec:segmentation}.

\subsection{Why is Random Placement not Working?}\label{sec:prelim}
As we discovered in the Section~\ref{sec:random_copy-paste}, random copy-paste
data augmentation does not bring improvement when used to augment object
detection datasets. There are multiple possible reasons for observing this
behavior, such as violation of context constraints imposed by the dataset,
objects looking ``out of the scene'' due to different illumination conditions or
simply artifacts introduced due to blending techniques.
To investigate this phenomenon, we conduct a study, that aims to better
understand (i) the importance of visual context for object detection, (ii) the
role of illumination conditions and (iii) the impact of blending artefacts.
For simplicity, we choose the first 5 categories of VOC'12, namely \textit{aeroplane,
bike, bird, boat, bottle}, and train independent detectors per category.

\textit{Baseline when no object is in context.} 
To confirm the negative influence of random placing, we consider one-category
detection, where only objects of one selected class are annotated with bounding
boxes and everything else is considered as background. Images that do not
contain objects of the selected category become background images. After
training 5 independent detectors as a baseline, we construct a similar experiment
by learning on the same number of instances, but considering as positive
examples only objects that have been synthetically placed in a random context.
This is achieved by removing from the training data all the images that have an
object from the category we want to model, and replacing it by an instance of
this object placed on a background image. The main motivation for such study is
to consider the extreme case where (i) no object is placed in the right context;
(iii) all objects may suffer from rendering artefacts. As shown in
Table~\ref{tab:preliminary}, the average precision degrades significantly by
about $14\%$ compared to the baseline. As a conclusion, either visual context is
indeed crucial for learning, or blending artefacts is also a critical issue. The
purpose of the next experiment is to clarify this ambiguity.

\begin{table}[btp!] 
\centering
\renewcommand{\arraystretch}{1.4}
\renewcommand{\tabcolsep}{0.5mm}
\begin{tabular}{l| c c c c c |c}
  Method         & aero & bike & bird & boat & bottle & average \\
  \hline
  Base-DA              & 58.8 & 64.3 & 48.8 & 47.8 & 33.9   & 48.7\\
  Random-DA            & 60.2 & 66.5 & 55.1 & 41.9 & 29.7   & 48.3\\
  \hline
  Removing context     & 44.0 & 46.8 & 42.0 & 20.9 & 15.5   & 33.9\\
  \hline
  Enlarge + Reblend-DA & 60.1 & 63.4 & 51.6 & 48.0 & 34.8   & 51.6\\
\end{tabular}
\caption{Ablation study on the first five categories of VOC'12. All models are
  learned independently. We compare classical data augmentation techniques
  (Base-DA), approaches obtained by copy-pasting objects, either randomly
  (Random-DA) or by preserving context (Enlarge+Reblend-DA). The line ``Removing context''
  corresponds to the first experiment described in Section~\ref{sec:prelim};
  Enlarge-Reblend corresponds to the second experiment.}
\label{tab:preliminary}
\end{table}

\begin{figure*}[hbtp!]
% \vspace*{-0.8cm}
\begin{center}
  \includegraphics[width=0.99\linewidth,trim=60 110 120 60,clip]{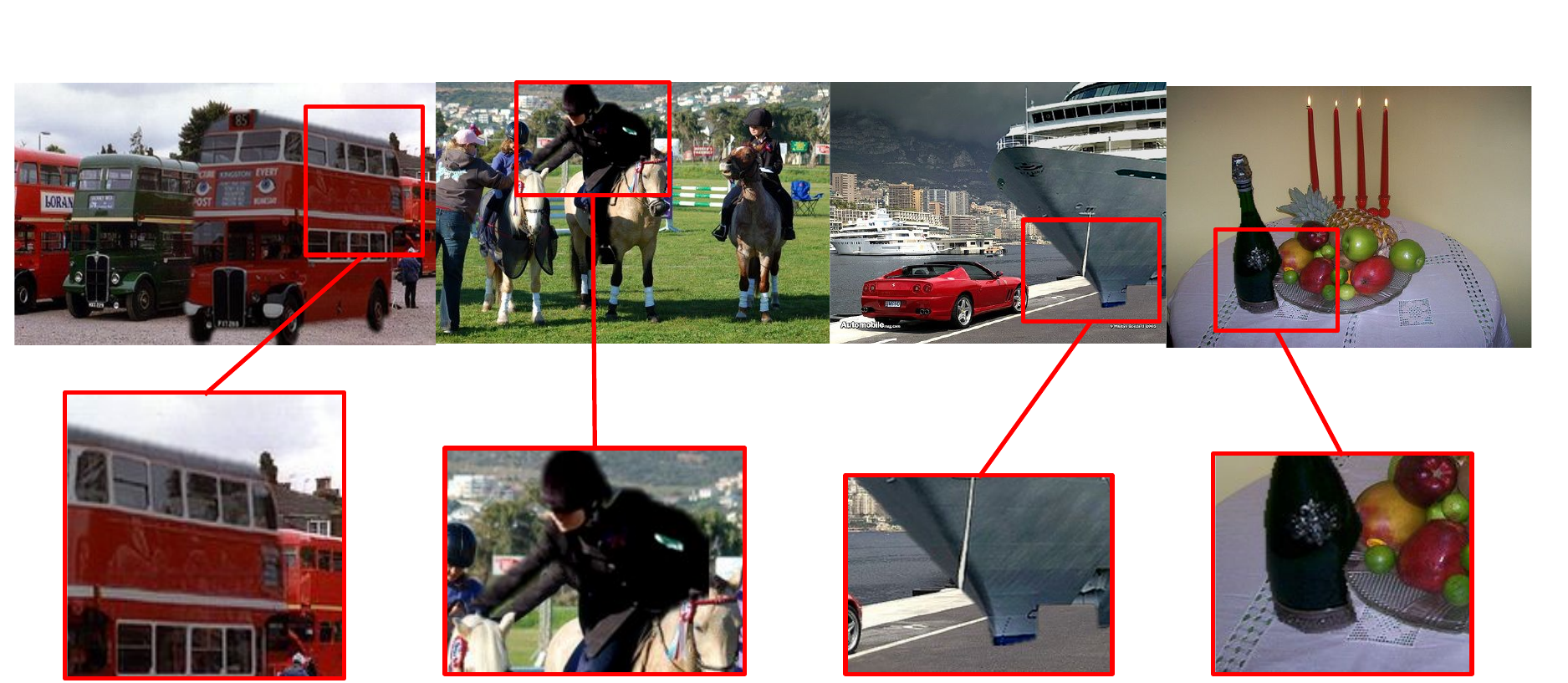}
\end{center}
\caption{\textbf{Illustration of artifacts arising from enlargement augmentation.}
  In the enlargement data augmentation, an instance is cut out of the image,
  up-scaled by a small factor and placed back at the same location. This approach 
  leads to blending artefacts.  Modified images are given in the top row.
  Zoomed parts of the images centered on blending artifacts are presented in the
  bottom line.}
\label{fig:enlargement}
\end{figure*}

\textit{Impact of blending when the context is right.} 
In the previous experiment, we have shown that the lack of visual context and
the presence of blending artefacts may explain the performance drop observed
in the third row of Table~\ref{tab:preliminary}. Here, we propose a simple experiment
showing that neither (iii) blending artefacts nor (ii) illumination difference
are critical when objects are placed in the right context: the experiment
consists of extracting each object instance from the dataset, up-scale it by a
random factor slightly greater than one (in the interval $[1.2,1.5]$), and blend
it back at the same location, such that it covers the original instance. To
mimic the illumination change we apply a slight color transformation to the
segmented object. As a result, the new dataset benefits slightly from data
augmentation (thanks to object enlargement), but it also suffers from blending
artefacts for \emph{all object instances}. As shown on the forth row of
Table~\ref{tab:preliminary}, this approach improves over the baseline, which
suggests that the lack of visual context is probably the key explaining the
result observed before. The experiment also confirms that the presence of
difference in illumination and blending artefacts is not critical for the object
detection task. Visual examples of such artefacts are presented in
Figure~\ref{fig:enlargement}.

\begin{table*}[hbtp!] 
\centering
\renewcommand{\arraystretch}{1.3}
\renewcommand{\tabcolsep}{0.5mm}
\resizebox{\linewidth}{!}{
\begin{tabular}{l|c c c c c c c c c c c c c c c c c c c c |c|}
method      & aero & bike & bird & boat & bott. & bus & car  & cat  & chair & cow & table & dog & horse & mbike & pers. & plant & sheep & sofa & train & tv   & avg.\\
  \hline
Base-DA     & 58.8 & 64.3 & 48.8 & 47.8 & 33.9 & 66.5 & 69.7 & 68.0 & 40.4 & 59.0 & 61.0  & 56.2 & 72.1 & 64.2  & 66.7  & 36.6  & 54.5  & 53.0 & 73.4  & 63.6 & 58.0\\

Random-DA   & 60.2 & 66.5 & 55.1 & 41.9 & 29.7 & 66.5 & 70.0 & 70.1 & 37.4 & 57.4 & 45.3  & 56.7 & 68.3 & 66.1  & 67.0  & 37.0  & 49.9  & 55.8 & 72.1  & 62.6 & 56.9\\

Enlarge-DA  & 60.1 & 63.4 & 51.6 & 48.0 & 34.8 & 68.8 & 72.1 & 70.4 & 41.1 & 63.7 & 62.3  & 56.3 & 70.1 & 67.8  & 65.3  & 37.9  & 58.1  & 61.2 & 75.5  & 65.9 & 59.7\\

Context-DA  & 68.9 & 73.1 & 62.5 & 57.6 & 38.9 & 72.5 & 74.8 & 77.2 & 42.9 & 69.7 & 59.5  & 63.9 & 76.1 & 70.2  & 69.2  & 43.9  & 58.3  & 59.7 & 77.2  & 64.8 & 64.0\\
  \hline
Impr. Cont. & \bf{10.1} & \bf{8.7}  & \bf{13.7}  & \bf{9.2}  & 5.0  & 6.0  & 5.1  & \bf{9.2}  & 2.5  & \bf{10.7} & 1.5   & \bf{7.5} &  4.0  & 6.0   & 2.5   & \bf{7.3}   & 3.8   & 6.7  & 4.2   & 1.2  & 5.8 \\
\end{tabular}
}
\caption{Comparison of detection accuracy on \texttt{VOC07-test} for the
  single-category experiment. The models are trained independently for each
  category, by using the $1\,464$ images from \texttt{VOC12train-seg}. The first
  row represents the baseline experiment that uses standard data augmentation
  techniques. The second row uses in addition copy-pasting of objects with random
  placements. ``Enlarge-DA'' augmentation blends up-scaled instances back in
  their initial location, which is given in row 3. The forth row presents the
  results achieved by our context-driven approach and the last row presents the
  improvement it brings over the baseline. The numbers represent average
  precision per class in \%. Large improvements over the baseline (greater than
  $7\%$) are in bold. All numbers are averaged over 3 independent experiments.
}
\label{tab:single_cat}
\end{table*}

\begin{table*}[hbtp!] 
\centering
\renewcommand{\arraystretch}{1.3}
\renewcommand{\tabcolsep}{0.5mm}
\resizebox{\linewidth}{!}{
\begin{tabular}{l| c |c c c c c c c c c c c c c c c c c c c c | c |}
model & CDA  & aero & bike & bird & boat & bott. & bus & car & cat & chair & cow & table & dog & horse & mbike & pers. & plant & sheep & sofa & train & tv & avg.\\
  \hline
  \multirow{2}{*}{BlitzNet300} &            & 63.6 & 73.3 & 63.2 & 57.0 & 31.5 & 76.0 & 71.5 & 79.9 & 40.0 & 71.6 & 61.4 & 74.6 & 80.9 & 70.4 & 67.9 & 36.5 & 64.9 & 63.0 & 79.3 & 64.7 & 64.6\\
                               & \checkmark  & 69.9 & 73.8 & 63.9 & 62.6 & 35.3 & 78.3 & 73.5 & 80.6 & 42.8 & 73.8 & 62.7 & 74.5 & 81.1 & 73.2 & 68.9 & 38.1 & 67.8 & 64.3 & 79.3 & 66.1 & \bf{66.5}\\
  \hline
  \multirow{2}{*}{F-RCNN} &            & 65.8 & 70.9 & 66.5 & 54.6 & 45.9 & 72.7 & 72.9 & 80.3 & 36.8 & 70.3 & 48.0 & 78.9 & 70.7 & 70.6 & 66.3 & 33.1 & 64.7 & 59.8 & 71.8 & 61.1 & 63.1\\
                          & \checkmark & 67.4 & 67.7 & 64.9 & 58.0 & 50.4 & 71.6 & 74.9 & 80.4 & 36.8 & 70.2 & 56.4 & 75.7 & 73.7 & 71.6 & 71.5 & 39.4 & 68.6 & 63.5 & 67.7 & 60.1 & \bf{64.5} \\
\end{tabular}
}
\caption{Comparison of detection accuracy on \texttt{VOC07-test} for the
  multiple-category experiment. The model is trained on all categories at the same
  time, by using the $1\,464$ images from \texttt{VOC12train-seg}.
  The first column specifies the detector used in the experiment, the second
  column notes if Context-driven Data Augmentation (CDA) was used. The numbers
  represent average precision per class in \%.}
\label{tab:multiple}
\end{table*}

\subsection{Object Detection Augmentation on VOC PASCAL}\label{sec:det_voc}
In this subsection, we are conducting experiments on object detection by
augmenting the PASCAL VOC'12 dataset. In order to measure the impact of the
proposed technique in a ``small data regime'', we pick the single-category
detection scenario and also consider a more standard multi-category setting. We test
single-shot region-based families of detectors---with BlitzNet and Faster-RCNN
respectively---and observe improved performance in both cases.

\subsubsection{Single-category Object Detection}\label{sec:single}
In this section, we conduct an experiment to better understand the effect of the
proposed data augmentation approach, dubbed ``Context-DA'' in the different
tables, when compared to a baseline with random object placement ``Random-DA'',
and when compared to standard data augmentation techniques called ``Base-DA''.
The study is conducted in a single-category setting, where detectors are trained
independently for each object category, resulting in a relatively small number
of positive training examples per class. This allows us to evaluate the
importance of context when few labeled samples are available and see if
conclusions drawn for a category easily generalize to other ones.

The baseline with random object placements on random backgrounds is conducted in
a similar fashion as our context-driven approach, by following the strategy
described in the previous section. For each category, we treat all images with
no object from this category as background images, and consider a collection of
cut instances as discussed in Section~\ref{sec:datasets}. During training, we
augment a negative (background) image with probability 0.5 by pasting up to two
instances on it, either at randomly selected locations (Random-DA), or using our
context model in the selected bounding boxes with top scores (Context-DA). The
instances are re-scaled by a random factor in $[0.5, 1.5]$ and blended into an
image using a randomly selected blending method mentioned in
Section~\ref{sec:datasets}. For all models, we train the object detection
network for 6K iterations and decrease the learning rate after 2K and 4K
iterations by a factor 10 each time. The context model was trained in
``small-data regime'' for 2K iterations and the learning rate was dropped once
after 1.5K steps. The results for this experiment are presented in
Table~\ref{tab:single_cat}.

The conclusions are the following: random placement indeed hurts the
performance on average. Only the category bird seems to benefit significantly
from it, perhaps because birds tend to appear in various contexts in this
dataset and some categories significantly suffer from random placement such as
boat, table, and sheep.  Importantly, the visual context model always improves
upon the random placement one, on average by 7\%, and upon the baseline that uses
only classical data augmentation, on average by 6\%. Interestingly, we identify
categories for which visual context is crucial (aeroplane, bird, boat, bus, cat,
cow, dog, plant), for which context-driven data augmentation brings more than
7\% improvement and some categories that display no significant gain or losses
(chair, table, persons, tv), where the difference with the baseline is less
noticeable (around 1-3\%).

\begin{table*}[hbtp!] 
\centering
\renewcommand{\arraystretch}{1.3}
\renewcommand{\tabcolsep}{0.5mm}
\resizebox{\linewidth}{!}{
\begin{tabular}{l|c c c c c c c c c c c c c c c c c c c c | c |}
method  & aero & bike & bird & boat & bott. & bus & car & cat & chair & cow & table & dog & horse & mbike & pers. & plant & sheep & sofa & train & tv & avg.\\
  \hline
  Base-DA     & 79.0 & 43.7 & 65.8 & 57.9 & 53.8 & 83.8 & 77.9 & 76.7 & 19.2 & 56.6 & 46.6 & 67.6 & 59.0 & 73.1 & 77.9 & 46.8 & 69.4 & 37.8 & 73.7 & 70.3 & 63.3\\

  Random-DA   & 78.1 & 47.1 & 75.4 & 57.8 & 57.2 & 83.5 & 76.2 & 76.6 & 20.5 & 57.0 & 43.1 & 69.2 & 57.5 & 71.5 & 78.2 & 40.0 & 63.3 & 42.0 & 74.5 & 64.1 & 63.1 \\

  Enlarge-DA  & 77.2 & 45.4 & 67.9 & 57.9 & 61.0 & 84.1 & 78.8 & 76.3 & 20.3 & 58.4 & 46.9 & 67.5 & 60.5 & 73.9 & 78.1 & 45.2 & 71.1 & 38.8 & 73.6 & 71.1 & 64.1 \\

  Context-DA  & 81.7 & 46.4 & 73.4 & 60.7 & 59.4 & 85.3 & 78.8 & 79.1 & 20.6 & 60.0 & 48.0 & 68.1 & 62.2 & 75.3 & 78.8 & 47.6 & 71.6 & 39.9 & 73.6 & 70.3 & 65.4\\
  \hline
  Impr. Cont. & \textbf{2.7} & \textbf{2.7} & \textbf{7.6} & \textbf{2.8} & \textbf{4.6} & 1.5 & 1.1 & 2.3 & 1.4 & \textbf{3.4} & 1.4 & 0.5 & \textbf{3.2} & 2.3 & 2.2 & 0.9 & 0.8 & 2.1 & -0.1 & 0 & 2.1 \\
\end{tabular}
}
\vspace*{0.1cm}
\caption{Comparison of segmentation accuracy on \texttt{VOC12val-seg}. The model
  is trained on all 20 categories by using the $1\,464$ images from
  \texttt{VOC12train-seg}. Base-DA represents the baseline experiment that uses
  standard data augmentation techniques. Context-DA uses also our context-driven
  data augmentation. Random-DA is its context-agnostic analogue. Enlarge-DA
  corresponds to randomly enlarging an instance and blending it back. The last row
  presents absolute improvement over the baseline. The numbers represent IoU per
  class in \%. Categories enjoying an improvement higher than $2.5\%$ are in bold.
  All numbers are averaged over 3 independent experiments. }
\label{tab:seg}
\end{table*}

\begin{table}[hbtp!]
\centering
\renewcommand{\arraystretch}{1.2}
\renewcommand{\tabcolsep}{1.0mm}
\resizebox{\linewidth}{!}{
  \begin{tabular}{l | c | c c c | c c c}
    { Model} & DA &{@0.5:0.95} & {@0.5} &  @0.75 & S & M & L \\
    \hline
    \multicolumn{8}{c}{Object Detection} \\
    \hline
    BlitzNet300   &      &  27.3  & 46.0 & 28.1 & 10.7 & 26.8 & 46.0 \\
    BlitzNet300   & Rnd  &  26.8  & 45.0 & 27.6 & 9.3 & 26.0 & 45.7 \\
    BlitzNet300   & Cont &  \bf{28.0}  & \bf{46.7} & \bf{28.9} & 10.7 & \bf{27.8} & \bf{47.0} \\

    \hline
    \red{Mask-RCNN}    &       &      38.6  &     59.7 &     42.0   &     22.1 &      41.5  &     50.6  \\
    \red{Mask-RCNN}    & Rnd   &      36.9  &     57.3 &     39.7   &     20.5 &      39.6  &     48.0  \\
    \red{Mask-RCNN}    & Cont  &  \bf{39.1} & \bf{60.3} & \bf{42.3} & \bf{22.4} & \bf{42.2} & \bf{51.2} \\
    \hline
    \multicolumn{8}{c}{Instance Segmentation} \\
    \hline
    \red{Mask-RCNN}    &       &      34.5  &     56.5 &      36.3 &      15.7 &      37.1 &      52.1  \\
    \red{Mask-RCNN}    & Rnd   &      33.6  &     55.2 &      35.8 &      14.8 &      35.5 &      50.0  \\
    \red{Mask-RCNN}    & Cont  &  \bf{34.8} & \bf{57.0} & \bf{36.5} & \bf{15.9} & \bf{37.6} & \bf{52.5} \\
  \end{tabular}
  }
  \caption{Comparison of object detection and instance segmentation accuracy on
    \texttt{COCO-val2017} for the multiple-category experiment. The model is trained
    on all categories at the same time, by using the $118\,783$ images from
    \texttt{COCO-train2017}. The first column specifies a model used to solve a
    task, the second column notes if Context-driven (Cont) or random-placement
    (Rnd) Data Augmentation was used. For different IoU thresholds @0.5:0.95, @0.5
    and @0.75) and for different object size (S, M, L), the numbers represent mAP in
    \%. Best results are in bold. }
  \label{tab:coco}
\end{table}

\subsubsection{Multiple-Categories Object Detection}\label{sec:multiple}
In this section, we conduct the same experiment as in Section~\ref{sec:single},
but we train a single multiple-category object detector instead of independent
ones per category. Network parameters are trained with more labeled data (on
average 20 times more than for models learned in Table~\ref{tab:single_cat}).
When training the context model, we follow the ``normal-data strategy''
described in Section \ref{sec:details} and train the model for 8K iterations,
decreasing the learning rate after 6K steps. The results are presented in
Table~\ref{tab:multiple} and show a modest average improvement of $2.1\%$ for a
single shot and $1.4\%$ for a region-based detector on average over the corresponding
baselines, which is relatively consistent across categories. This confirms
that data augmentation is crucial when few labeled examples are available.

\begin{table*}[!t]
\centering
\renewcommand{\arraystretch}{1.3}
\renewcommand{\tabcolsep}{0.5mm}
\resizebox{\linewidth}{!}{
% \begin{tabular}{l|c c c c c c c c c c}
\begin{tabular}{l|c c c c c c c c c c c c c c c c c c c c |c|}
  Aug. type & aero & bike & bird & boat & bottle & bus  & car  & cat  & chair & cow  & table & dog  & horse & mbike & pers. & plant & sheep & sofa & train & tv   & avg.\\
  \hline
  Inst seg  & \bf{68.9} & \bf{73.1} & \bf{62.5} & \bf{57.6} & \bf{38.9}   & \bf{72.5} & \bf{74.8} & \bf{77.2} & \bf{42.9}  & \bf{69.7} & 59.5  & \bf{63.9} & 76.1  & 70.2  & \bf{69.2}  & \bf{43.9}  & 58.3  & \bf{59.7} & 77.2  & 64.8 & \bf{64.0} \\
  Gt seg    & 67.8 & 70.3 & 61.5 & 56.6 & 38.2   & 71.2 & 74.7 & 75.7 & 41.6  & 68.3 & 59.0  & 63.2 & 75.6  & 71.0  & 68.7  & 42.6  & 59.5  & 59.1 & \bf{78.4}  & \bf{65.3} & 63.4 \\
  Weak seg  & 68.9 & 71.3 & 59.0 & 54.2 & 37.3   & 71.9 & 74.5 & 75.2 & 40.8  & 67.6 & \bf{59.8}  & 62.8 & \bf{76.4}  & \bf{71.3}  & 68.4  & 43.8  & \bf{59.9}  & 57.2 & 76.6  & 64.4 & 63.0 \\
  \hline
  No        & 58.8 & 64.3 & 48.8 & 47.8 & 33.9 & 66.5 & 69.7 & 68.0  &  40.4 & 59.0  & 61.0       & 56.2 & 72.1 & 64.2 & 66.7      & 36.6 & 54.5 & 53.0  & 73.4  & 63.6 & 58.0\\
  \end{tabular}
}
\caption{Comparison of detection accuracy on \texttt{VOC07-test} for the
  single-category experiment. The models are trained independently on each
  category, by using the \texttt{VOC12train-seg}. The first column specifies
  the type of object mask used for augmentation: ground-truth instance
  segmentations (Inst. Seg.), ground-truth semantic segmentation (GT Seg.), or
  weakly-supervised semantic segmentations (Weak Seg.). Inst. seg. stands for the
  original instance segmentation ground truth masks. The numbers represent AP per
  class in \%. The best result for a category is in bold. All numbers are averaged
  over 3 independent experiments. }
\label{tab:single_weak}
\end{table*}

\begin{table*}[!t]
\centering
\renewcommand{\arraystretch}{1.3}
\renewcommand{\tabcolsep}{0.5mm}
\resizebox{\linewidth}{!}{
\begin{tabular}{l|c c c c c c c c c c c c c c c c c c c c |c|}
Aug. type & aero & bike & bird & boat & bott. & bus & car & cat & chair & cow & table & dog & horse & mbike & pers. & plant & sheep & sofa & train & tv & avg.\\
  \hline
Inst. seg  & \bf{69.9} & 73.8 & \bf{63.9} & \bf{62.6} & 35.3 & \bf{78.3} & 73.5 & 80.6 & \bf{42.8} & \bf{73.8} & 62.7 & \bf{74.5} & 81.1 & 73.2 & 68.9 & 38.1 & \bf{67.8} & 64.3 & 79.3 & \bf{66.1} & \bf{66.5}\\
GT Seg.    & 68.7 & 74.5 & 60.1 & 60.0 & 34.9 & 75.4 & \bf{74.4} & \bf{81.7} & 41.1 & 72.4 & \bf{64.2} & 74.4 & \bf{81.3} & \bf{74.6} & \bf{69.6} & \bf{39.7} & 67.6 & 64.2 & \bf{80.4} & 65.5 & 66.2\\
Weak Seg.  & 69.2 & \bf{75.2} & 63.2 & 59.8 & \bf{35.6} & 77.1 & 73.4 & 78.7 & 41.3 & 72.9 & 62.8 & 72.7 & 79.6 & 72.5 & 68.1 & 39.2 & 67.6 & \bf{66.1} & 79.5 & 64.2 & 65.9\\
  \hline
No         & 63.6 & 73.3 & 63.2 & 57.0 & 31.5 & 76.0 & 71.5 & 79.9 & 40.0 & 71.6 & 61.4 & 74.6 & 80.9 & 70.4 & 67.9 & 36.5 & 64.9 & 63.0 & 79.3 & 64.7 & 64.6\\
\end{tabular}
}
\caption{Comparison of detection accuracy on \texttt{VOC07-test} for the
  multy-category experiment depending on the type of object masks used for
  augmentation. The models are trained on all categories together, by using the
  $1\,464$ images from \texttt{VOC12train-seg}. The first column specifies the
  type of object mask used for augmentation: ground-truth instance segmentations
  (Inst. Seg.), ground-truth semantic segmentation (GT Seg.), or weakly-supervised
  semantic segmentations (Weak Seg.). Inst. seg. stands for the original instance
  segmentation ground truth masks. The numbers represent AP per class in \%. The
  best result for a category is in bold. All numbers are averaged over 3
  independent experiments. }
\label{tab:multy_weak}
\end{table*}

% \subsubsection{Multiple-Categories Object Detection on COCO}
% To check how our data augmentation strategy scales with a bigger dataset, we use
% COCO whose training set size is almost by 2 orders of magnitude larger than the
% previously used \texttt{voc12train-seg}. By design, the experiment is identical
% to the one presented in Section~\ref{sec:multiple}. However, for the COCO
% dataset we need to train a new context model. This is done by training for 350K
% iterations (decay at 250K) as described in Section~\ref{sec:details}. The non
% data-augmented baseline was trained according to \cite{blitznet}; when using our
% augmentation pipeline, we train the detector for 700K iterations and decrease
% the learning rate by a factor of 10 after 500K and 600K iterations.
% Table~\ref{tab:coco} shows that we are able to achieve a modest improvement of
% $0.7\%$, and that data augmentation still works and does not degrade the
% performance regardless the large amount of data available for training
% initially.

\subsection{Object Detection and Instance Segmentation Augmentation on COCO}\label{exp:coco}
In order to test our augmentation technique at large scale, we use in this
section the COCO dataset~\cite{coco} whose training set size is by 2 orders of
magnitude larger than \texttt{voc12train-seg}, and consider both object
detection and instance segmentation tasks.

\subsubsection{Object Detection with BlitzNet}\label{exp:coco_det}
By design, the experiment is identical to the one presented in
Section~\ref{sec:multiple}. However, for the COCO dataset we need to train a new
context model. This is done by training for 350K iterations (decay at 250K) as
described in Section~\ref{sec:details}. The non data-augmented baseline was
trained according to \cite{blitznet}; when using our augmentation pipeline, we
train the detector for 700K iterations and decrease the learning rate by a
factor of 10 after 500K and 600K iterations. Table~\ref{tab:coco} shows that we
are able to achieve a modest improvement of $0.7\%$, and that data augmentation
still works and does not degrade the performance regardless the large amount of
data available for training initially.

\subsubsection{Detection and Segmentation with Mask-RCNN}\label{exp:coco_seg}
\red{For this experiment, we use Mask-RCNN~\cite{mask_rcnn} that jointly solves
object detection and instance segmentation. When training the baseline model, we
closely follow original
guidelines\footnote{\url{https://github.com/facebookresearch/Detectron/blob/master/MODEL_ZOO.md}}
and train the model with 2x schedule (for 180K iterations) to maximize the
baseline's performance. Training the model with 1x schedule (for 90K iterations)
results in underfitting, while training with x4 schedule (for 360K iterations),
results in overfitting. In order to improve the performance of Mask-RCNN for both tasks, we
train the model with x4 schedule and use the context-driven data augmentation.
In order to reduce pasting artifacts negatively affecting
Mask-RCNN, we decrease the augmentation probability during the training. More
precisely, augmentation probability is set to 0.5 in the beginning of the
training and then linearly decreased to 0 by the end of the training procedure. Training
with constant augmentation probability did not improve the performance over the
x2 baseline. 
% We believe that, often non-realistic images and sometimes missing
% instance annotations, arising when copy-pasting nested instances, are confusing
% for a state-of-the-art vision system. However, gradually shifting the training
% from diverse and noisy to limited and accurate annotations helped with
%     generalization. Apart from that,
On the other hand, gradually reducing augmentation probability results in less
aggressive regularization and brings more benefits when training on a large
dataset, such as COCO.
As Table~\ref{tab:coco} shows, following this
augmentation strategy results in a $0.5\%$ an $0.3\%$ mAP improvement for
detection and segmentation respectively, when comparing to the most accurate
baseline Mask-RCNN, trained with x2 schedule. Augmenting the training data
with random placement strategy hurts the performance substantially,
which highlights the importance of context for data augmentation.}

\subsection{Semantic Segmentation Augmentation}\label{sec:segmentation}
In this section, we demonstrate the benefits of the proposed data augmentation
technique for the task of semantic segmentation by using the VOC'12 dataset.
First, we set up the baseline by training the BlitzNet300~\cite{blitznet}
architecture for semantic segmentation. Standard augmentation techniques such as
flipping, cropping, color transformations and adding random noise were applied during the training,
as described in the original paper. We use \texttt{voc12train-seg} subset for
learning the model parameters. Following the training procedure described in
Section~\ref{sec:details}, we train the model for 12K iterations starting from
the learning rate of $10^{-4}$ and decreasing it twice by the factor of 10, after
7K and 10K steps respectively. Next, we perform data augmentation of the
training set with the proposed context-driven strategy and train the same
model for 15K iterations, dropping the learning rate at 8K and 12K steps. In
order to blend new objects in and to augment the ground truth we follow routines
described in Section~\ref{subsec:context_da}. We also carry out an experiment
where new instances are placed at random locations, which represents a
context-agnostic counterpart of our method. We summarize the results of all 3
experiments in Table~\ref{tab:seg}. As we can see from the table, performing
copy-paste augmentation at random locations for semantic segmentation slightly
degrades the model's performance by $0.2\%$. However when objects are placed in
the right context, we achieve a boost of $2.1\%$ in mean intersection over union.
These results resemble the case of object detection a lot and therefore highlight
the role of context in scene understanding. We further analyze the categories that benefit from our data
augmentation technique more than the others. If improvement for a class AP over
the baseline is higher than $2.5\%$, Table~\ref{tab:seg} marks the result in
bold. Again, we can notice correlation with the detection results from
Section~\ref{sec:single} which demonstrates the importance of context for the
categories that benefit from our augmentation strategy in both cases.

\subsection{Reducing the need for pixel-wise object annotation}\label{exp:weak}
Our data augmentation technique requires instance-level segmentations, which are
not always available in realistic scenarios. In this section, we relax the
annotation requirements for our approach and show that it is possible to use the
method when only bounding boxes are available.

\begin{figure}[btp!]
\begin{center}
  \includegraphics[width=0.99\linewidth,trim=35 50 30 35,clip]{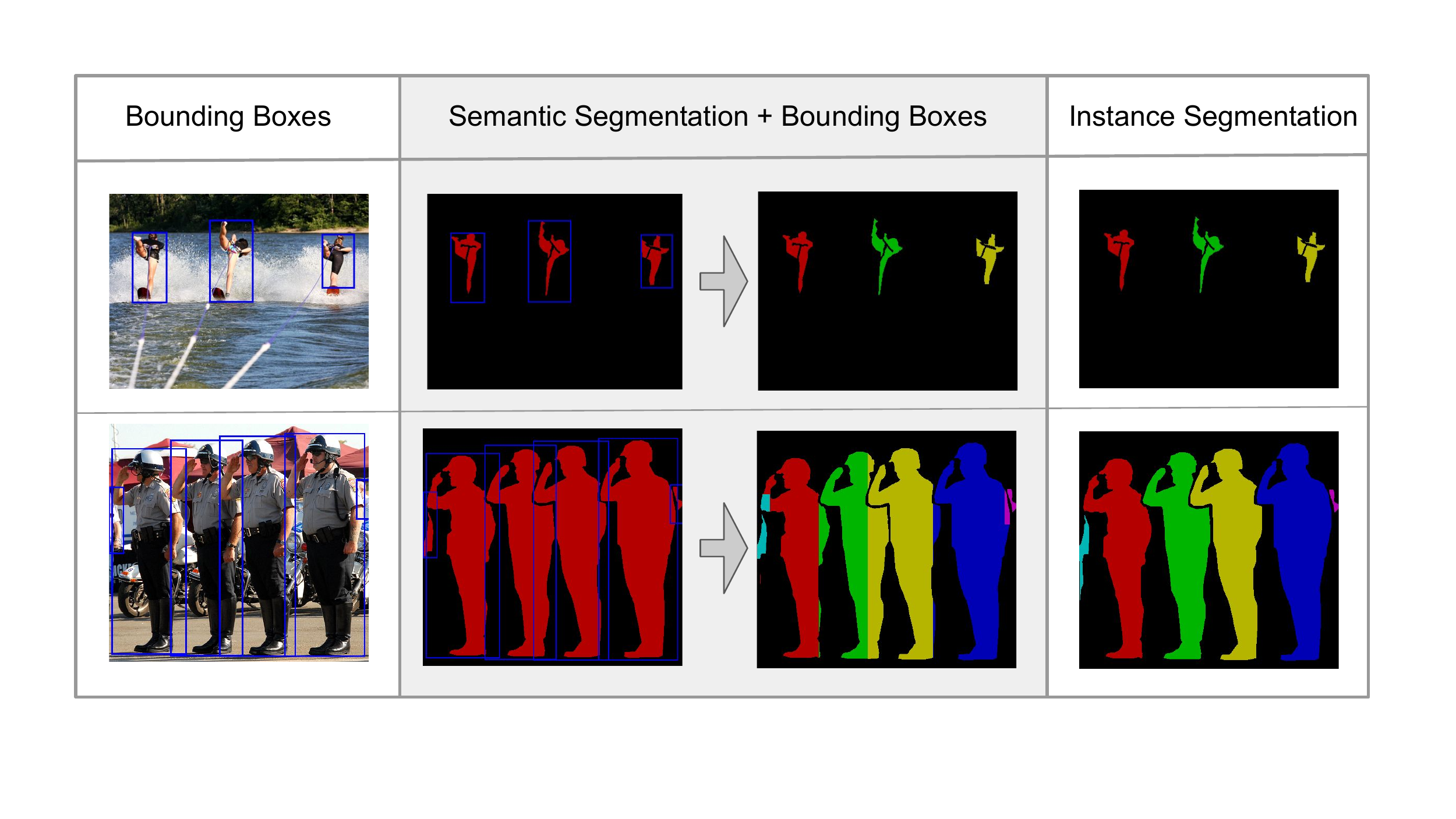}
\end{center}
\caption{\textbf{Possible types of instance-level annotation.}
  The left column presents an image annotated with object boxes. Column 2 shows
  semantic segmentation annotations with object boxes on top and approximate
  instance segmentations derived from it. The last column presents the original
  instance segmentation annotations.}
\label{fig:annotation_types}
\end{figure}

\textbf{Semantic segmentation + bounding box annotations.}
Instance segmentation masks provide annotations to each pixel in an image and
specify (i) an instance a pixel belongs to and (ii) class of that instance. If
these annotations are not available, one may approximate them with semantic
segmentation and bounding boxes annotations. Figure~\ref{fig:annotation_types}
illustrates possible annotation types and the difference between them. Semantic
segmentation annotations are also pixel-wise, however they annotate each pixel
only with the object category. Instance-specific information could be
obtained from object bounding boxes, however this type of annotation is not
pixel-wise and in some cases is not sufficient to assign each pixel to the
correct instance. As Figure~\ref{fig:annotation_types} suggests, as long as
a pixel in semantic map is covered by only one bounding box, it uniquely defines
the object it belongs to (row 1); otherwise, if more than one box covers the
pixel, it is not clear which object it comes from (row 2). When deriving
approximate instance masks from semantic segmentation and bounding boxes (see
Figure~\ref{fig:annotation_types}, column 2), we randomly order the boxes and
assign pixels from a semantic map to the corresponding instances. Whenever a
pixel could be assigned to multiple boxes we choose a box that comes first in the
ordering. Once the procedure for obtaining object masks is established we are
back to the initial setting and follow the proposed data augmentation routines
described above. As could be seen in
Tables~\ref{tab:single_weak}~and~\ref{tab:multy_weak} detection performance
expiriences a slight drop of $0.6\%$ in single-category and $0.3\%$ in
multi-category settings respectively, comparing to using instance segmentation
masks. These results are promising and encourage us to explore less elaborate
annotations for data augmentation.

\textbf{Bounding box annotations only.}
Since we have an established procedure for performing data augmentation with
semantic segmentation and bounding boxes annotations,
the next step to reducing pixel-wise annotation is to approximate segmentation
masks. We employ weakly-supervised learning to estimate segmentations
from bounding boxes and use the work of \cite{khoreva2017simple}. When trained
on the \texttt{VOC12train} dataset, augmented with more training examples
according to \cite{khoreva2017simple,gould2009decomposing}, it achieves $65.7\%$
mIoU on the
\texttt{VOC12val-set}.
Unfortunately, we have found that naively applying this solution for estimating
segmentation masks and using them for augmentation results in worse performance.
The reason for that was low quality of estimated masks. First, inaccurate
object boundaries result in non-realistic instances and may introduce biases in
the augmented dataset. But more importantly, confusion between classes may hampers the
performance. For example, augmenting a category ``cow''
with examples of a ``sheep'' class may hurt the learning process.
Hence, we need a model with a more discriminative classifier. To this end we propose the
following modifications to the segmentation method: we change the architecture from
DeepLab\_v1 \cite{chen2014semantic} to DeepLab\_v4 \cite{chen2018deeplab},
perform multi-scale inference and process the resulting masks with a conditional random field. The
later helps to refine the object edges, which was found not necessary in the
original work of \cite{chen2018deeplab}, when learning with full supervision. By
training on the same data as the original method of \cite{khoreva2017simple} but
with the proposed modifications we achieve $75.8\%$
mIoU, which is more than 10\% improvement to the initial pipeline. This accuracy
seems to be sufficient to use automatically-estimated segmentation masks for augmentation
purposes.

When the semantic maps are estimated, we follow the augmentation routines of the
previous section with only one difference; specifically, an instance is kept if the
bounding box of its segmentation covers at least $40\%$ of its corresponding
ground truth box. Otherwise, the object is not used for data augmentation. The results of applying this
strategy to the single- and multy-category object detection are presented in
Table~\ref{tab:single_weak}~and~\ref{tab:multy_weak}, respectively. 
Table~\ref{tab:single_weak} shows which categories are unable to provide high-quality
masks, even though the quality seems to be sufficient to improve upon the non-augmented baseline. It is surprising
that by using object
boxes instead of segmentation masks we lose only $0.6\%$ of mAP in the multi-class
scenario while still outperforming non-augmented training by $1.6\%$. These results
show that the method is widely applicable even in the absence of segmentation
annotations.

\subsection{Studying the Importance of Context Modeling Quality for Scene
Understanding}\label{exp:context_quality}
First, we make an assumption that the quality of a context model is
mainly influenced by the amount of data it has received for training. Hence, to
study this relation, we mine a bigger dataset
\texttt{VOC07-trainval+VOC12-trainval} which results in 16551 images. Then, we
proceed by taking subsets of this dataset of increasing size and train the
context model on them. Finally, we use the obtained context models to augment
\texttt{VOC12-trainval} and train BlitzNet300 on it for detection and segmentation.
Table \ref{tab:context_size} summarizes the object detection performance on
\texttt{VOC07-test} and semantic segmentation performance on
\texttt{VOC12val-seg}.
In the current experiment, 10\% of the full set (1655 images) is roughly equal
to the size of the \texttt{VOC12train-seg} (1464 images) initially used for
training the context model. As we increase the data size used for context modeling, we
can see how both detection and segmentation improve; however, this gain diminishes as the data size keeps growing. This
probably mean that to improve scene understanding, the context model has to
get visual context ``approximately right'' and further improvement is most
likely limited by other factors such as unrealistic generated scenes and
limited number of instances that are being copy-pasted. On the other hand,
if the context model is trained with little data, as
in the case of using only 5\% of the full set, our augmentation strategy tends
to the random one and shows little improvement.

\begin{table}[btp!] 
\centering
\renewcommand{\arraystretch}{1.4}
\renewcommand{\tabcolsep}{1.5mm}
\resizebox{\linewidth}{!}{
\begin{tabular}{l |c| c c c c c c}
  $\%$ of data used  & 0 & 5 & 10 & 25 & 50 & 75 & 100 \\
  \hline
  Det. mAP        & 64.6  & 65.3 & 66.1 & 66.4 & 66.7 & 66.9 & 66.9 \\
  Seg. mIoU       & 63.3  & 64.6 & 65.1 & 65.3 & 65.5 & 65.9 & 66.0 \\
  \hline
\end{tabular}
}
\caption{Object detection and semantic segmentation performance
  depending on amount of data used for building the context model.
  First row depicts the portion (in \%) of
  the \texttt{VOC0712trainval} used for training the
  context model. Second column corresponds to performance of baseline models.
  The second row gives the final detection mAP \% evaluated
  on \texttt{VOC07test}, while the third row lists segmentation mIou in \%
  on \texttt{VOC12val-seg}. For both tasks we used BlitzNet300 trained on
  augmented \texttt{VOC12train-seg}.}
\label{tab:context_size}
\end{table}

%% file: ccl.tex
In this paper, we introduce a data augmentation technique dedicated to scene
understanding problems. From a methodological point of view, we show that this
approach is effective and goes beyond traditional augmentation methods. One
of the keys to obtain significant improvements in terms of accuracy was to
introduce an appropriate context model which allows us to automatically find
realistic locations for objects, which can then be pasted and blended at in the
new scenes. While the role of explicit context modeling was so far unclear for
scene understanding, we show that it is in fact crucial when performing data
augmentation and learn with fewer labeled data, which is one of the major issues
deep learning models are facing today.